\documentclass[preprint]{IEEEtran}

\usepackage{microtype}
\usepackage{graphicx}
\usepackage{subcaption}
\usepackage{booktabs} % for professional tables
\usepackage{tabularx}
\usepackage{algorithm,bm,algorithmicx,algpseudocode}
\algnewcommand{\LineComment}[1]{\Statex \hspace{\algorithmicindent}$\triangleright$ #1}

\usepackage{cite}

\usepackage{amsmath}
\usepackage{amssymb}
\usepackage{mathtools}
\usepackage{amsthm}
\usepackage{comment}
\usepackage{dsfont}

\usepackage{hyperref}
\usepackage[capitalize,noabbrev]{cleveref}
\crefname{appendix}{App.}{Apps.}
\Crefname{appendix}{App.}{Apps.}
\crefname{figure}{Fig.}{Figs.} % match the IEEEtran caption style
\Crefname{figure}{Fig.}{Figs.}

\theoremstyle{plain}

\theoremstyle{definition}

\theoremstyle{remark}

\newcommand{\TIMECONSTANT}{\mu}
\newcommand{\DEV}{ \sigma}

\title{Encoding and Decoding Temporal Signals with Spiking Bandpass Wavelets}

\author{Jens~E.~Pedersen,~Tony~Lindeberg,~and~Peter~Gerstoft% <-this % stops a space
\thanks{J. E. Pedersen and P. Gerstoft are with the Department of Electrical and Photonics Engineering, Technical University of Denmark, Denmark (e-mail: jegpe@dtu.dk; pegers@dtu.dk).}% <-this % stops a space
\thanks{T. Lindeberg is with the Department of Computational Science and Technology, KTH Royal Institute of Technology, Sweden (e-mail: tony@kth.se).}
\thanks{Support from the Novo Nordisk Foundation (NNF24OC0089302) and the Swedish Research Council (2022-02969) is gratefully acknowledged.}
}

\begin{document}

\maketitle

\begin{abstract}
Spike-based encodings are sparse and energy-efficient. However, previous theory for such spiking representations is largely disconnected from most signal processing literature and does not extend to multi-scale frames.
We recast spike-encoders as time-causal overcomplete wavelet frames with closed-form bandwidth tiling and reconstruction error bounds.
We show that scale covariance, frame bounds, and reconstruction guarantees hold for any finite-energy input in both encoding and decoding steps.
The proposed spiking bandpass wavelets preserve the sparsity and locality of spiking representations: A threshold parameter sets the event rate, and the reconstruction error grows linearly with it, up to spike quantization and time discretization.
On ECG and audio datasets, we reach reconstruction accuracy comparable to continuous wavelet transforms, multi-channel integrate-and-fire time encoding machines, and Sigma-Delta modulation.
The experiments demonstrate robustness of the proposed spiking bandpass wavelets to noise and spike-time jitter, and that the spike activity shifts across channels as predicted by the covariance property under temporal rescaling.
Every operation reduces to a leaky integrate-and-fire difference with thresholding, and both the encoder and the decoder map directly to existing neuromorphic hardware.
\end{abstract}


\section{Introduction}

\begin{figure*}[t]
    \centering
    \includegraphics[width=0.85\textwidth]{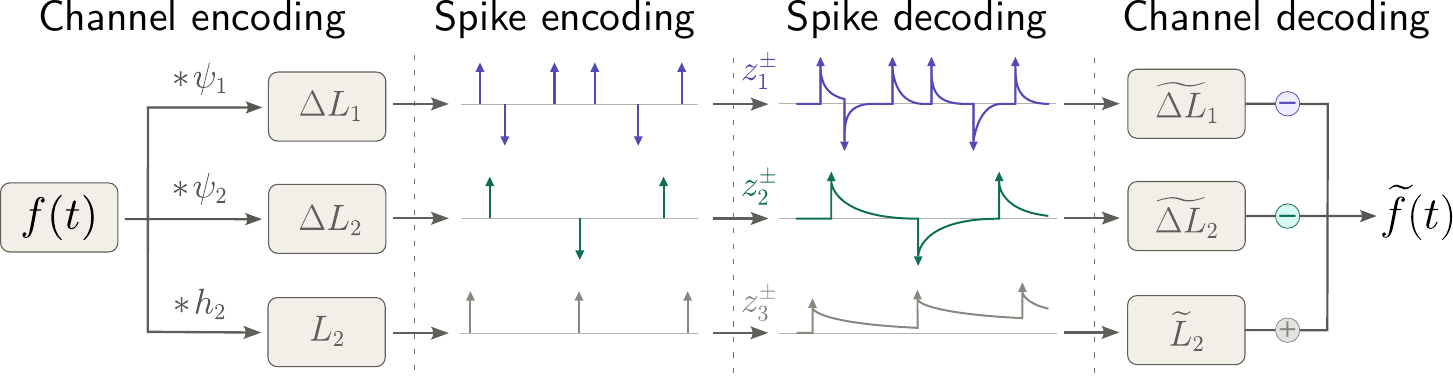}
    \caption{Proposed \cref{alg:encoderdecoder}, demonstrated with three channels.
        A signal $f(t)$ is encoded into three channels: two bandpass representations $\Delta L$ by wavelets $\psi$ \eqref{eq:difference-of-reconstruction} and a lowpass representation $L$ with smoothing kernel $h$ \eqref{eq:scale-space}.
        The channels are then encoded as sparse polarity spikes $z_k^\pm(t)$ \eqref{eq:spike_train}.
        To recover an approximation of the original signal $f(t)$, the three channels are recreated from the spikes \eqref{eq:channel_reconstruction} and combined to form $\widetilde{f}(t)$ \eqref{eq:per_spike_decoupled_reconstruction}.
    }
    \label{fig:spiking_autoencoder}
\end{figure*}

Analog-to-digital conversion (ADC) underpins every digital signal processing pipeline.
Uniform Nyquist-Shannon sampling \cite[Thm. 1]{shannon1949communication} is a well-understood approach, but it is agnostic to signal structure: it indiscriminately samples both silence and activity.

Biological sensory systems sample differently.
Rather than measuring at a fixed rate, spikes are emitted only when the input changes, producing sparse, event-driven, and highly energy-efficient representations \cite{roy2019spikebased}.
However, their energy advantage over dense activations is not automatic, but depends on the event rate and hardware \cite{dampfhoffer2023are}.
A leaky integrate-and-fire (LIF) neuron implements a time-causal convolution \eqref{eq:lif_srm} followed by a threshold comparator, encoding change as spikes rather than amplitude as samples.
Neuromorphic hardware strives to exploit this efficiency, but a principled signal-processing framework for spike-based encoding is missing.

The encoded representations should be (1) localized in frequency, (2) scale covariant, to enable structurally similar processing over different temporal scales, and (3) equipped with formal bounds that guarantee stable reconstruction regardless of the input.
Wavelet filter banks satisfy all three properties by construction \cite{daubechies1992ten} and are the natural mathematical home for a principled spike-based ADC.
A spiking analogue is missing: a wavelet frame whose coefficients are emitted as spike events rather than continuous amplitudes.
Such a frame would inherit the covariant, multi-resolution structure of classical wavelet theory, while producing sparse, event-driven output.

Neuroscience-driven models are typically presented with information-theoretic \cite{warland1996spikes} or probabilistic \cite{gerstner2014neuronal} foundations rather than signal processing guarantees.
Recent work explores the integrate-and-fire model from a signal processing perspective \cite{mose2024integrate}, but does not extend to multi-scale frames.
Sigma-delta ($\Sigma\Delta$) modulators \cite{yoon2017lif} offer recovery guarantees but are single-channel or restricted to shift-invariant subspaces.
Time encoding machines \cite{adam2020sampling} operate on fixed shift-invariant subspaces rather than scale-parameterized families.
Quantization in overcomplete frame expansions is well studied in its own right \cite{goyal2001quantized}, but for fixed frames rather than for a covariant, scale-parameterized family generated by a spiking encoder.
Prior work \cite{pedersen2025covariant, pedersen2026scale} established covariance guarantees for the LIF response and sketched a spiking wavelet construction.
We expand this with two time-causal wavelet families realized as spiking analysis and synthesis operators: difference-of-truncated-exponentials (DoE), built from a single leaky integrator per scale, and difference-of-time-causal-limit-kernels (DoT), using cascades of leaky integrators (\cref{fig:spiking_autoencoder}).
The spiking wavelets rely on bandpass wavelets \cite{lindeberg2025time, pedersen2027frame}.
But where spiking autoencoders learn their encodings from data \cite{comsa2021spiking,kamata2022fully,fang2026eventdriven}, both families here exist in closed form, so their bandwidth and reconstruction bounds follow analytically, and we can derive reconstruction guarantees, error bounds, and quantitative evaluation.

Our contributions are threefold: \\
\textbf{(i)~Spiking wavelets:} two spiking time-causal scale-covariant wavelet families, based on difference-of-truncated-exponentials (DoE) \eqref{eq:doe_wavelet} and difference-of-time-causal-limit-kernels (DoT) \eqref{eq:dot_wavelet}.\\
\textbf{(ii)~Guarantees:} a closed-form reconstruction error bound \eqref{eq:total_error_bound} for the causal spiking wavelets: unconditionally for the DoT, and under mild assumptions on the signal spectrum for the DoE.
This extends Time Encoding Machines \cite{lazar2004perfect, gontier2014sampling, adam2020sampling} to overcomplete multi-scale frames.\\
\textbf{(iii)~Experiments:} reconstruction on ECG signals (MIT-BIH \cite{moody2001impact}) and audio (LibriSpeech \cite{panayotov2015librispeech}) reaching reconstruction RMSE comparable to classical and continuous wavelets, and to the MC-IFTEM and $\Sigma\Delta$ event encoders at a matched spike rate.
Additionally, the DoE wavelets export to existing neuromorphic backends via the Neuromorphic Intermediate Representation \cite{pedersen2024neuromorphic}, enabling hardware-backed execution in the future.

\section{Background}

This section covers scale-space theory, including time-causal scale-space representations, wavelets, wavelet frames, and spiking neuron models.

\subsection{Scale spaces}
Scale-space theory parameterizes a signal $f\colon \mathbb{R} \to \mathbb{R}$ over a scale parameter $\DEV \in \mathbb{R}_+$ into the scale-space representation $L\colon \mathbb{R} \times \mathbb{R}_+ \to \mathbb{R}$ \cite{koenderink1984structure}.
Computing the scale-space representation $L$ corresponds to convolving the signal $f(t)$ with smoothing kernels $h(t;\ \DEV)$ \cite[(1.2)]{lindeberg1994scalespace}
\begin{align} \label{eq:scale-space}
    L(t;\ \DEV) = h(t;\ \DEV) * f(t)
    = \int_{\xi = -\infty}^{\infty} h(\xi;\ \DEV) \, f(t - \xi) \, {\rm d}\xi,
\end{align}
where "$*$" denotes convolution, with the zero-scale condition
\begin{equation} \label{eq:zero-scale-representation}
    L(t;\; \DEV = 0) = f(t) \qquad \text{thus}\quad  h(t;\; \DEV = 0) = \delta(t).
\end{equation}
Scale-space kernels $h$ are required to diminish variations over scales and to form a continuous semi-group under convolution, leaving the Gaussian with mean $0$ and standard deviation $\DEV$ as the unique admissible kernel over spatial domains
\begin{equation} \label{eq:gaussian}
    h_{\rm Gauss}(t;\; \DEV) = \frac{1}{\sqrt{2 \pi}\DEV}e^{-\frac{t^2}{2\DEV^2}}.
\end{equation}
The Gaussian kernel is physically interpreted as the Green's function of the heat equation, over the variance $\DEV^2$ \cite[p.~370]{koenderink1984structure}.
Scale-space representations are scale covariant \cite{lindeberg2023time}, or scale equivariant.
Consider the scaling transformation $t' = s \, t$ and the corresponding scale parameter transformation $\DEV' = s \, \DEV$ for any scaling factor $s > 0$.
A scale-covariant representation satisfies
\begin{equation}
    L'(t';\; \DEV') = L(t;\; \DEV),
\end{equation}
meaning that the scale-space representation $L'$ over the transformed domain captures the same structural information as $L$, but observed at a different scale.

Representing a signal at multiple scales requires a set of scale levels $\DEV_k$, $k \in \mathbb{N}$, between some minimum scale $\DEV_0$ and maximum scale $\DEV_K$.
A logarithmic spread of scales retain self-similarity \cite{lindeberg2023time} and describe temporal relations in memory \cite{howard2025learning}.
Denote the ratio between adjacent scale levels as $c$, with the default choices $c \in \{\sqrt{2}, 2\}$:
\begin{equation} \label{eq:scale_level_spread}
    \DEV_{k} = c \, \DEV_{k-1}.
\end{equation}
We write $L(t;\; \DEV_k, c)$ for the scale-space representation \eqref{eq:scale-space} at the $k$-th level of this grid, $c$ denoting the distance to neighbouring scales.

\subsubsection{Time-causal scale spaces} \label{sec:time-causal_scale_spaces}
For time-causal operations, a continuous semi-group structure is not possible: the causality constraint requires one-sided kernels which cannot generate a continuous semi-group of smoothing kernels while retaining variation-diminishing properties \cite{pedersen2027frame, lindeberg2023time}.
The scale parameter must be discretized with the unique variation-diminishing kernel with one-sided support, the {\it truncated exponential} \cite{schoenberg1948variationdiminishing}
\begin{equation} \label{eq:truncated_exponential}
    h_{\rm exp}(t;\; \TIMECONSTANT) = \begin{cases}
        \TIMECONSTANT^{-1} \exp{(-t / \TIMECONSTANT)} & t > 0          \\
        0                                             & t \leqslant 0,
    \end{cases}
\end{equation}
where $\TIMECONSTANT$ is the time constant and also the delay.
Time-causal scale spaces are built by cascading $K$ such kernels with geometrically spaced time constants as \eqref{eq:scale_level_spread}.
% The ratio $c > 1$  is the ratio between the scale values as adjacent scale levels, as measured in units of the standard deviation $\DEV$ of the composed temporal smoothing kernels.

\begin{figure}
    \centering
    \includegraphics[width=0.9\linewidth]{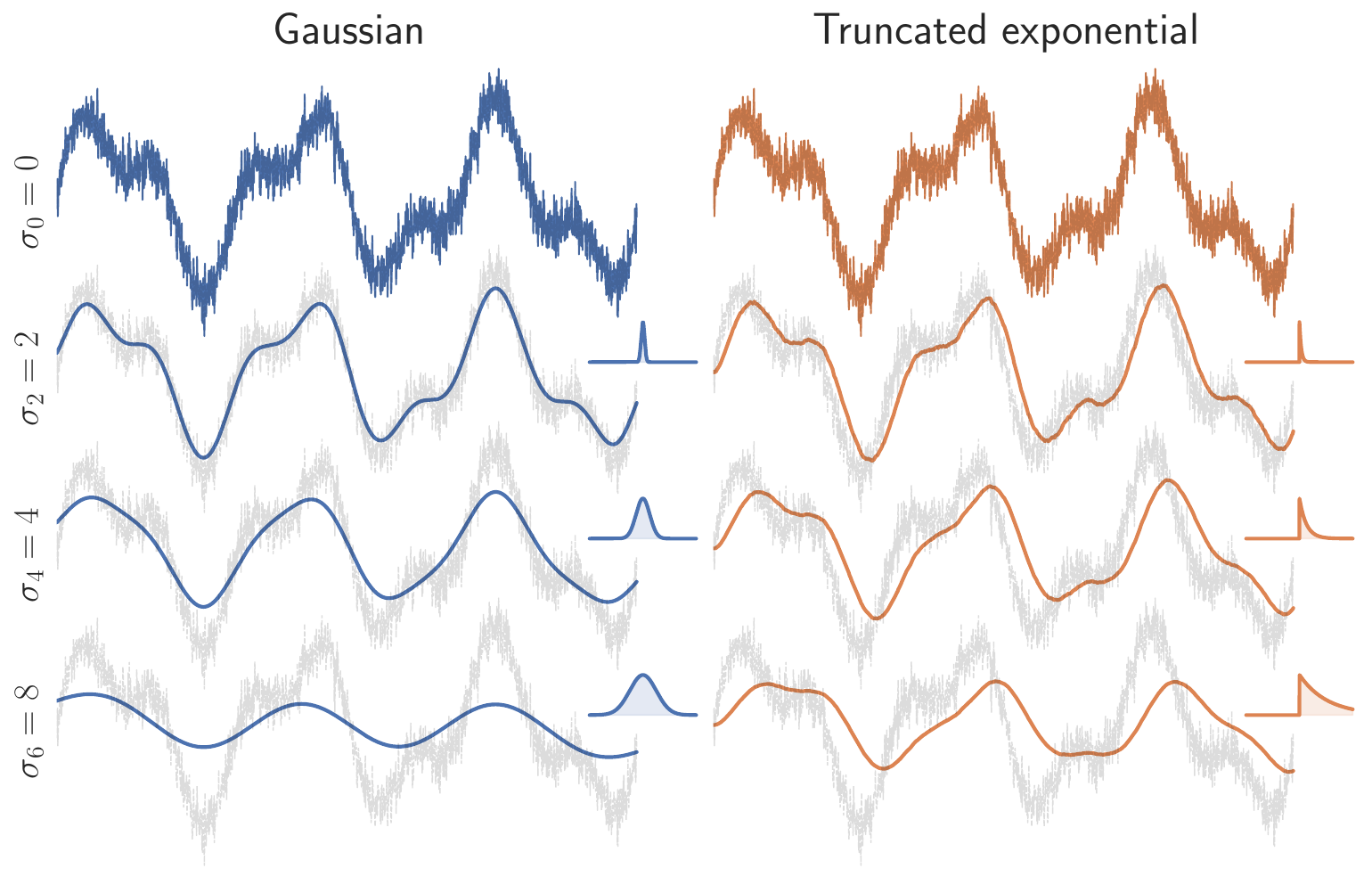}
    \caption{Scaled representations of a noisy signal smoothed by Gaussian \eqref{eq:gaussian} and truncated exponential \eqref{eq:truncated_exponential} kernels, shown in the small insets.
        The original signal (top row) at scale $\DEV=0 $ \eqref{eq:zero-scale-representation} smoothens as $\DEV$ or $\TIMECONSTANT$ increases.
    }
    \label{fig:scale_smoothing}
\end{figure}

\Cref{fig:scale_smoothing} illustrates the scale representation at varying temporal scales using the Gaussian \eqref{eq:gaussian} and truncated exponential \eqref{eq:truncated_exponential} kernels.
The causal representation thus has temporal delays that increase with the temporal scale, see \cref{fig:scale_smoothing} right.

\subsection{Wavelets and frames} \label{sec:wavelets_frames}
Wavelets decompose a signal $f(t)$ into scale and shift components using the wavelet generating function
\begin{equation}
    \psi(t; a, b) = |a|^{-1/2} \, \psi\left((t - b)/a\right)
\end{equation}
with scale $a$ and temporal shift $b$.
The wavelet transform is defined as
\begin{equation} \label{eq:wavelet_transform}
    \begin{split}
        \big(T(f)\big)(t; a, b) & = \langle f(t), \psi(t; a, b) \rangle                                                              \\
                                & = \int_{-\infty}^{\infty} f(t) \, |a|^{-1/2} \, \overline{\psi \left((t - b)/a \right)}\ {\rm d}t,
    \end{split}
\end{equation}
where $\langle \cdot, \cdot \rangle$ denotes the inner $L^2$ product, and $\overline{\ \psi \ }$ denotes the complex conjugate of $\psi$ \cite[p.~24]{daubechies1992ten}.

For $\psi$ to serve as a mother wavelet, it must satisfy the admissibility criterion \cite[(2.4.1)]{daubechies1992ten}
\begin{equation} \label{eq:wavelet_admissibility}
    C_\psi = \int_0^\infty \frac{|\hat{\psi}(\omega)|^2}{\omega} \, \mathrm{d}\omega < \infty,
\end{equation}
where $\hat{\psi}$ is the Fourier transform of $\psi$, which implies the zero-mean condition \cite[(1.2.3)]{daubechies1992ten}
\begin{equation} \label{eq:zero_mean}
    \int_{-\infty}^\infty \psi(t) \, {\rm d}t = 0.
\end{equation}

\subsubsection{Frame operator}
Frames are possibly overcomplete, continuous-time expansions.
A family of wavelets $\psi_k$, indexed by $k \in K$ scales, is a frame if there exist bounds $A$ and $B$, where $0 < A \leqslant B < \infty$, such that for all $f$ in a Hilbert space \cite[(3.1.2)]{daubechies1992ten}
\begin{equation} \label{eq:frame_bounds}
    A \, \|f\|^2 \leqslant \sum_{k \in K} |\langle f, \psi_k\rangle |^2 \leqslant B \, \|f\|^2.
\end{equation}
When $A = B$, the frame is tight, providing uniform representations across scales.
When $A < B$, the frame is overcomplete, offering robustness, but at the cost of redundant parameters.
For a filterbank of shift-invariant channels, the bounds are given by the energy capture function \cite[Thm.~5.11]{mallat2009wavelet}
\vspace{-0.1em}
\begin{equation} \label{eq:energy_capture_def}
    S(\omega) = \sum_{k \in K} |\hat{\psi}_k(\omega)|^2,
    \vspace{-0.2em}
\end{equation}
with $A = \inf_\omega S(\omega)$ and $B = \sup_\omega S(\omega)$.
A frame operator $\Lambda$ transforms a continuous-time signal from $L_2(\mathbb{R}) \to \ell_2(K)$ as $\Lambda(f(t))_{k \in K} = \langle f(t), \psi_k\rangle$.
Since $\Lambda$ is bounded, the Hermitian adjoint is well defined $\Lambda^*: \ell_2(K) \to L_2(\mathbb{R})$
\begin{equation} \label{eq:lambda_adjoint}
    \Lambda^*\big(\Lambda(f(t))\big) = \sum_k \langle f(t), \psi_k\rangle \, \psi_k.
\end{equation}
This we use to define a dual family of vectors $\tilde{\psi}_k \in L_2(\mathbb{R})$
\begin{equation} \label{eq:dual_frame_operator}
    \tilde{\psi}_k = \left(\Lambda^*\Lambda\right)^{-1} \psi_k.
\end{equation}
$\tilde{\psi}_k$ is itself a frame with bounds $B^{-1}$ and $A^{-1}$ \eqref{eq:frame_bounds} \cite[(3.2.6)]{daubechies1992ten},
permitting signal reconstruction \cite[(3.2.8)]{daubechies1992ten}
\begin{equation} \label{eq:frame_reconstruction}
    f(t) = \sum_{k \in K} \langle f, \psi_k\rangle \, \tilde{\psi}_k.
\end{equation}

\subsection{Spiking neuron models}
\begin{figure}
    \centering
    \includegraphics[width=\linewidth]{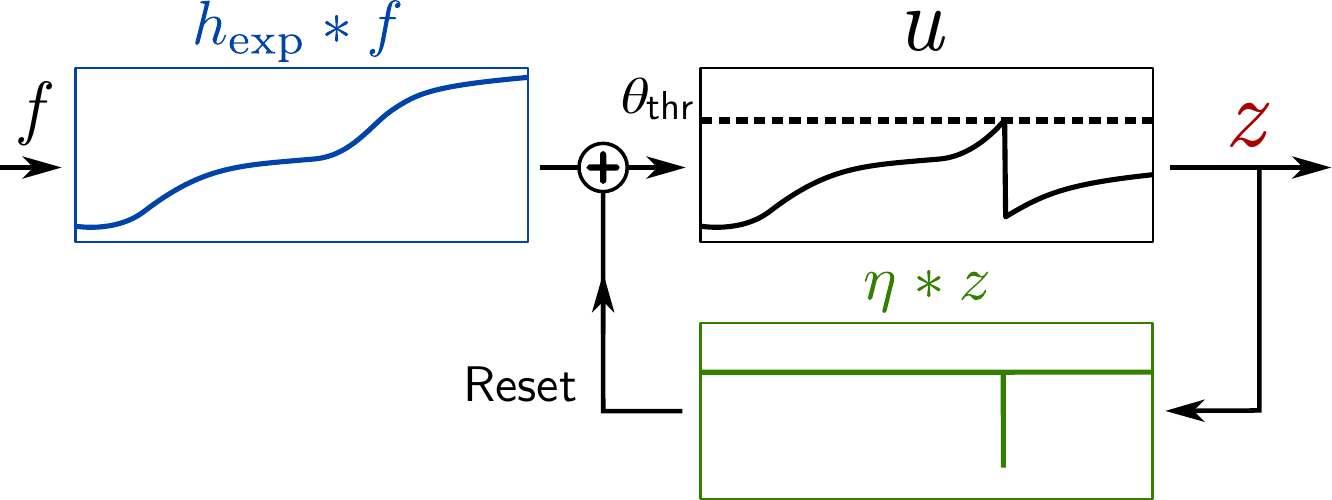}
    \caption{The leaky integrate-and-fire (LIF) model \eqref{eq:lif_srm} as a composition of linear systems.
        An integrator $h_{\rm exp}$ convolves a signal $f$, captured by a voltage membrane $u$ that outputs spikes ($z$) when $u \geqslant \theta_{\text{thr}}$.
        % The reset kernel $\eta = -\theta_{\rm thr}$ subtracts $\theta_{\rm thr}$ from $u$ following a spike.
    }
    \label{fig:lif_srm}
\end{figure}
Physical neurons implement $h_{\rm exp}$ \eqref{eq:truncated_exponential} directly \cite[(1.6)]{gerstner2014neuronal}.
The membrane voltage $u$ of a leaky integrator dissipates according to the RC circuit derived from Kirchhoff's law, $C \dot{u} = -(u - u_\mathrm{rest})/R + I(t)$, where $C$ is the membrane capacitance, $R$ the membrane resistance, $u_\mathrm{rest}$ the resting potential, and $I(t)$ the input current in amperes.
Measuring $u$ relative to $u_\mathrm{rest}$, scaling the input current to units of voltage as $f(t) = R I(t)$, and identifying the membrane time constant $\TIMECONSTANT = R C$ with the scale parameter of \eqref{eq:truncated_exponential} gives
\begin{equation} \label{eq:li_dot}
    \TIMECONSTANT \, \dot{u} = -u + f(t).
\end{equation}
The impulse response of \eqref{eq:li_dot} is exactly $h_{\rm exp}(t; \TIMECONSTANT)$, so for arbitrary input $f(t)$ the membrane voltage implements the time-causal scale-space convolution \cite[(22)]{pedersen2025covariant} $u(t;\TIMECONSTANT) = (h_{\rm exp}(\cdot; \TIMECONSTANT) * f)(t)$.
In biological neurons, the continuous membrane voltage $u$ is communicated to downstream neurons via discrete spike events.
A spike is emitted when $u$ hits a threshold $\theta_{\rm thr}$, after which $u$ resets.
The spike train $z(t)$ is a sum of Dirac impulses at firing time $t_{\rm spike} \in \mathcal{S}$, where $\mathcal{S}$ is the set of all firing times \cite[(1.14)]{gerstner2014neuronal}
\begin{align} \label{eq:spike_train}
    z(t) & = \sum_{t_\text{spike}}\delta(t - t_{\text{spike}}),
\end{align}
after which the membrane resets by a reset kernel $\eta$
\begin{equation} \label{eq:reset_kernel}
    u(t) = (\eta * z)(t).
\end{equation}
Combining the leaky integrator \eqref{eq:li_dot}, the spike function \eqref{eq:spike_train} (left), and the reset term \eqref{eq:reset_kernel}, we arrive at the LIF model (\cref{fig:lif_srm}) \cite[(6.28)]{gerstner2014neuronal}
\begin{align} \label{eq:lif_srm}
    u(t) & = (\eta*z)(t) + (h_{\rm exp} * f)(t) \nonumber                                                                                                                                                           \\
         & =\underbrace{-\theta_{\text{thr}}\ z(t)}_{\text{Reset integrand}} + \underbrace{\int_{\xi=0}^\infty f(t - \xi) \, \frac{1}{\TIMECONSTANT}e^{-\xi/\TIMECONSTANT} \, {\rm d}\xi}_{\text{Input integrand}}.
\end{align}
Defining the reset kernel $\eta(t) = -\theta_\mathrm{thr}\, \delta(t)$ feeds each spike back into the membrane, closing the loop in \cref{fig:lif_srm}.
Setting the smoothing kernel from \eqref{eq:scale-space} $h = h_\mathrm{exp}$, the membrane voltage is the time-causal scale-space representation, $u(t;\TIMECONSTANT) = L(t;\TIMECONSTANT)$, so the construct is scale covariant \cite[(24)]{pedersen2025covariant}.
Under temporal scaling operations $t' = s \, t$ and $\TIMECONSTANT' = s \, \TIMECONSTANT$ with $f'(t') = f(t)$, the substitution $\xi' = s \, \xi$ in \eqref{eq:lif_srm} leaves the reset and input terms unchanged, so
\begin{equation} \label{eq:srm_temporal_covariance}
    L'(t';\; \TIMECONSTANT') = L(t;\; \TIMECONSTANT).
\end{equation}
\Cref{tab:nomenclature} summarizes the symbols used in this paper along with their meaning and their initial appearance.
\begin{table}[h]
    \small
    \begin{tabularx}{\columnwidth}{@{}r >{\raggedright\arraybackslash}X l@{}}
        \toprule
        \textbf{Symbol}      & \textbf{Meaning}                                                   & \textbf{Appearance}                                                   \\
        \midrule
        {\Large $\hat{\phantom{a}}$}   & Fourier transform                                                  & \eqref{eq:wavelet_admissibility}                                      \\
        {\Large $\bar{\phantom{a}}$}   & Channel normalization ($\bar{x}_k$, $\bar\psi$); average ($\bar{r}$, $\bar{G}$) & \eqref{eq:scale_channel_norm} \eqref{eq:spike_rate}      \\
        {\Large $\tilde{\phantom{a}}$} & Reconstruction ($\widetilde{f}$, $\widetilde{x}_k$); dual ($\tilde{\psi}_k$) & \eqref{eq:per_spike_decoupled_reconstruction} \eqref{eq:dual_frame_operator} \\
        $A, B$               & Lower and upper frame bounds                                       & \eqref{eq:frame_bounds} \\
        $c$                  & Ratio between adjacent scale levels                                & \eqref{eq:scale_level_spread}                                         \\
        $f(t)$               & Real-valued signal at time $t$                                     & \eqref{eq:scale-space}                                                \\
        $\widetilde{f}(t)$   & Reconstructed signal at time $t$                                   & \eqref{eq:per_spike_decoupled_reconstruction}                         \\
        $h_\mathrm{Gauss}$   & Gaussian                                                           & \eqref{eq:gaussian}                       \\
        $h_\mathrm{exp}$     & Truncated exponential                                              & \eqref{eq:truncated_exponential}                       \\
        $h_\Psi$             & Time-causal limit kernel                                           & \eqref{eq:time-causal_limit_kernel_convolution}                       \\
        $K$                  & Number of bandpass channels. Scale levels run $\DEV_0$ (finest) to $\DEV_K$ (coarsest) & \eqref{eq:scale_level_spread}                                       \\
        $N$                  & Number of cascaded truncated exponentials in $h_\Psi$              & \eqref{eq:time-causal_limit_kernel_convolution}                       \\
        $\lvert f \rvert$    & Number of samples in $f(t)$                                        & \eqref{eq:spike_rate}                                                 \\
        $p_i$                & Polarity of spike $i$                                              & \eqref{eq:best_fit_A_spike}                                           \\
        $R_k$                & Spike-synthesis kernel of channel $k$                              & \eqref{eq:channel_reconstruction}                                     \\
        $S(\omega)$          & Energy capture function                                            & \eqref{eq:energy_capture_def}                                         \\
        $\theta_{\rm thr}$   & Firing threshold of the LIF quantizer                              & \eqref{eq:lif_srm}                                                    \\
        $\Delta t$           & Simulation time step                                               & \eqref{eq:channel_error_disc}                                         \\
        $\Omega_k$           & Essential bandwidth of channel $k$. The channel is not band-limited, but all of its spectral mass beyond $|\omega| > \Omega_k$ integrates to at most $\Delta_\Omega$ & \eqref{eq:bandwidth_tail_integral}                                        \\
        $L(t; \DEV_k, c)$    & Scale-space representation at time $t$, at the $k$-th level of the $c$-spaced grid & \eqref{eq:scale-space} \\
        $\TIMECONSTANT$      & Time constant, temporal scale resolution                           & \eqref{eq:truncated_exponential} \eqref{eq:time-causal_limit_kernel_convolution} \\
        $\Lambda$            & Frame operator                                                     & \eqref{eq:lambda_adjoint} \\
        $\DEV$               & Standard deviation, scale resolution                               & \eqref{eq:scale-space} \eqref{eq:scale_level_spread}                  \\
        $\mathcal{S}$        & The set of firing times in a spike train $z(t)$                    & \eqref{eq:spike_train}                                                \\
        $\psi_k$             & Bandpass wavelet of the channel at the $k$-th scale                & \eqref{eq:difference-of-reconstruction}                               \\
        $(\psi_k)_{k \in K}$ & Wavelet dictionary with $K$ scales                                 & \eqref{eq:wavelet_transform} \eqref{eq:scale_bandpass_reconstruction} \\
        $z(t)$               & A sum of Dirac impulses at firing times $\mathcal{S}$              & \eqref{eq:spike_train}                                                \\
        MC-IFTEM             & Multi-channel integrate-and-fire time-encoding machine (baseline)  & \eqref{eq:mc_iftem}                                                   \\
        $\Sigma\Delta$       & Sigma-delta modulator (baseline)                                   & \eqref{eq:sigma_delta}                                                \\
        \bottomrule
    \end{tabularx}
    \caption{Symbols and their meanings used throughout the paper with references to their first appearances.}
    \label{tab:nomenclature}
\end{table}

\section{Scale-covariant bandpass representations} \label{sec:causal_wavelets}
Sensory and physical signals carry features over multiple temporal scales.
To account for all scales in a spiking, or quantized, encoder, the representation must decompose into discrete scale channels, be covariant under temporal rescaling, respecting the self-similarity of multi-scale signals, and remove constant components so the channels admit wavelet frames; the latter making the frame bounds \cite{pedersen2027frame} possible.
Following the bandpass representation based on a Laplacian pyramid \cite{BA83-COM}, the notion of scale-space representation \cite{koenderink1984structure}, the multi-resolution decomposition of wavelet frames \cite{mallat1989theory}, and specifically the time-causal bandpass representation in \cite{lindeberg2025time},
%Marr-Hildreth zero-crossing detector \cite{marr1980theory}, 
we obtain all three by taking adjacent differences of scale-space representations: $L$ \eqref{eq:scale-space} is covariant by construction, and the difference channels ensure that the responses vanish at $\omega = 0$.
Taking adjacent differences at the geometric scale spacing \eqref{eq:scale_level_spread} yields a covariant representation,
\begin{multline} \label{eq:difference-of-reconstruction}
    \Delta L(t;\; \DEV_k, c) = L(t;\; \DEV_k, c) - L(t;\; \DEV_{k-1}, c) \\ = \psi(t;\; \DEV_k,c) * f(t),
\end{multline}
where $\psi$ is a bandpass kernel.
In the Gaussian case \eqref{eq:gaussian}, differences of adjacent scale levels approximate the Laplacian-of-Gaussian, the prototype bandpass filter \cite{marr1980theory}.

An exact reconstruction of a signal at scale $\DEV_k,$ is achieved by summing the representations along with a lowpass residual at the coarsest scale $\DEV_K$, seen by inserting \eqref{eq:difference-of-reconstruction} into \eqref{eq:scale-space},
\cite[(102)]{lindeberg2025time}
\begin{equation} \label{eq:scale_bandpass_reconstruction}
    \begin{split}
        \!\!\!L(t;\; \DEV_k, c) & \! =\! L(t;\; \DEV_K, c) -\! \sum_{j=k+1}^K \Delta L(t;\; \DEV_j, c)         \\
                                & \! =\! h(t;\; \DEV_K) * f(t) -\! \sum_{j=k+1}^K \psi(t;\; \DEV_j, c) * f(t).
    \end{split}
\end{equation}
Setting $\DEV_0 = 0$ gives $L(t; 0) = f(t)$ \eqref{eq:zero-scale-representation}, and we can reconstruct a wavelet frame \eqref{eq:frame_reconstruction} based on \eqref{eq:scale_bandpass_reconstruction} by summing over subsequent scales:
\cite[(103)]{lindeberg2025time}:
\begin{equation}
    \tilde{f}(t) = L(t; 0, c)
    %= L(t; \DEV_K, c) - \sum_{k=1}^K \Delta L(t; \DEV_k, c)
    = h(t;\; \DEV_K) * f(t) - \sum^K_{k=1} \psi(t;\; \DEV_k, c) * f(t).
\end{equation}
$K$ is the number of bandpass channels and the full filterbank has $K+1$ channels ($K$ bandpass and one lowpass residual at scale $\DEV_K$).
We construct an overcomplete frame, where the lowpass residual corresponds to the scaling function in a finite-level multi-resolution analysis \cite[(9)]{mallat1989theory}.

\subsection{Channel normalization}
Write $x_k$ for the response of channel $k$ (the bandpass response $\Delta L(t;\; \DEV_k, c)$ \eqref{eq:difference-of-reconstruction} for $k = 1, \dots, K$, and $x_{\rm lowpass}$ for the lowpass residual at scale $\DEV_K$).
The channel norm $\|\psi(t;\DEV_k, c)\|_2$ varies with the scale $\DEV_k$ and with the spacing $c$ \eqref{eq:scale_level_spread}, so we normalize each channel by its own norm
\begin{equation} \label{eq:scale_channel_norm}
    \bar{x}_k (t) = x_k (t) / \|\psi(t;\; \DEV_k, c)\|_2,
\end{equation}
as in the $L^2$ convention of the continuous wavelet transform, and likewise $\bar{x}_{\rm lowpass} = x_{\rm lowpass} / \|h(t;\DEV_K)\|_2$ for the lowpass residual.
The norm is a scalar, so this equals normalizing the kernel, $\bar\psi = \psi / \|\psi(t;\; \DEV_k, c)\|_2$ with $\bar{x}_k = \bar\psi * f$.
By Plancherel's theorem, the same scalar normalizes the channel in both the time and frequency domain
\begin{equation} \label{eq:scale_channel_norm_plancherel}
    \|\bar{\psi}(t;\; \DEV_k,c)\|_2^2 = \frac{1}{2\pi}\int |\hat{\bar\psi}(\omega;\; \DEV_k,c)|^2 \, \mathrm{d}\omega = 1,
\end{equation}
equalizing the energy of the channels, but not their peak gain.

\subsection{Bandpass wavelets}
We rely on three scale- and shift-parameterized bandpass kernels from \cite{lindeberg2025time,pedersen2027frame} as wavelets.

\subsubsection{Difference-of-Gaussians (DoG)}
The DoG wavelet
\begin{equation} \label{eq:dog_wavelet}
    \psi_{\rm DoG}(t;\; \DEV_k,c) = h_{\rm Gauss}(t;\; \DEV_k) - h_{\rm Gauss}(t;\; \DEV_{k-1}),
\end{equation}
has zero mean and is admissible according to \eqref{eq:wavelet_admissibility}.

\subsubsection{Difference-of-time-causal-limit-kernels (DoT)}
By convolving $h_\mathrm{exp}$ $N$ times, we construct a time-causal limit kernel with mean and standard deviation \cite[(8, 10)]{lindeberg2023time}
\begin{equation} \label{eq:time-causal_limit_kernel_convolution}
    \begin{aligned}
        h_\Psi(t;\; \DEV, c) & = *_{n=1}^N\ h_{\rm exp}(t;\; \TIMECONSTANT_n), \\
        \text{mean}[h_\Psi]  & = \sum_{n=1}^N \mu_n, \quad
        {\rm std}[h_\Psi] = \DEV = \sqrt{\sum_{n=1}^N \mu_n^2}.
    \end{aligned}
\end{equation}
When $N \to \infty$, we define bandpass filters at a specific scale $k$
\begin{equation} \label{eq:dot_wavelet}
    \psi_{\rm DoT}(t;\; \DEV_k, c) = h_\Psi(t;\; \DEV_k, c)  - h_\Psi(t;\; \DEV_{k - 1}, c).
\end{equation}

\subsubsection{Difference-of-truncated-exponentials (DoE)}
\begin{figure*}[t]
    \centering
    \includegraphics[width=\textwidth]{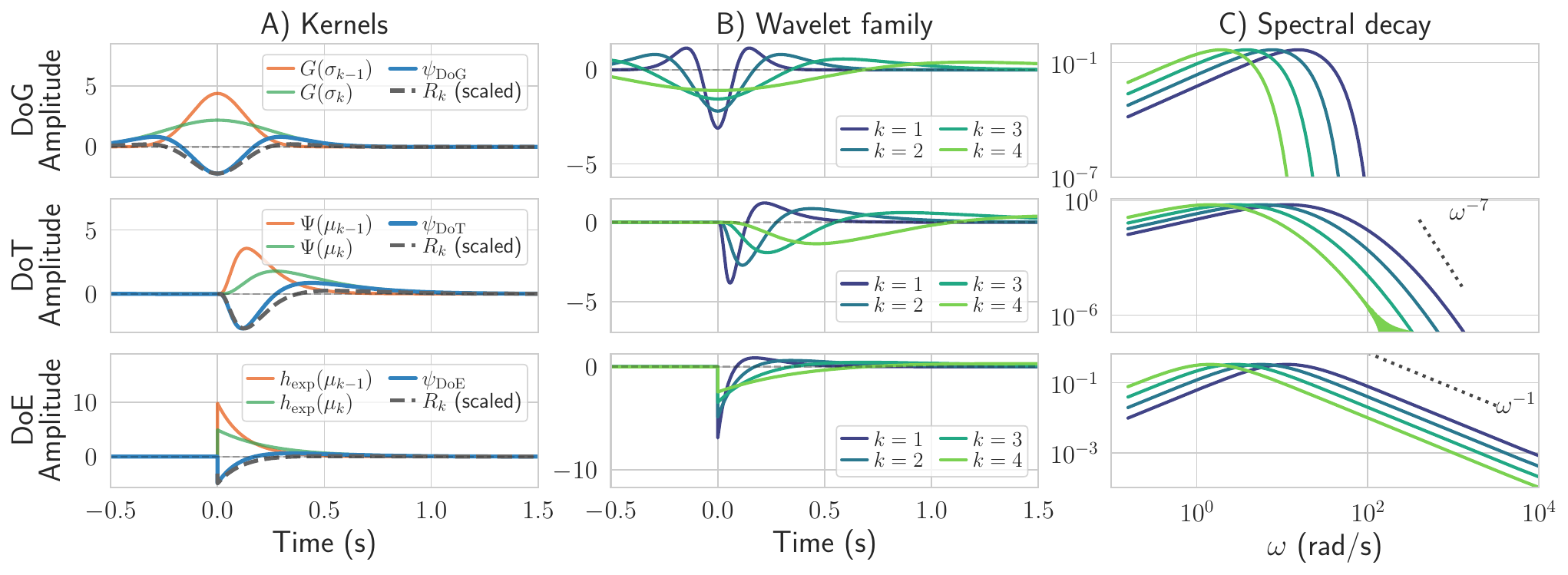}
    \caption{The three wavelet families at $c = 2$ over $K = 4$ bandpass channels.
        \textbf{A)} Each $\psi$ \eqref{eq:dog_wavelet}, \eqref{eq:dot_wavelet} ($N = 7$), \eqref{eq:doe_wavelet} as the difference of two adjacent smoothing kernels, normalized by the single scalar $\lVert\psi\rVert_2$ \eqref{eq:scale_channel_norm}, with the spike-synthesis kernel $R_k$ \eqref{eq:channel_reconstruction} for the same channel, matched to the peak of $\psi$, which it follows up to the delay.
        Broadening is a result of the membrane $h_{\rm exp}(\TIMECONSTANT_r)$.
        \textbf{B)} The normalized channels $\bar\psi_k$ \eqref{eq:scale_channel_norm} of each family.
        \textbf{C)} The same channels in the frequency domain with the asymptotes of \eqref{eq:spectral_decay} as dotted guides.
    }
    \label{fig:kernel_gallery}
\end{figure*}

For $N=1$,
\begin{equation} \label{eq:doe_wavelet}
    \psi_{\rm DoE}(t;\; \TIMECONSTANT_k, c) = h_{\rm exp}(t;\; \TIMECONSTANT_{k}) - h_{\rm exp}(t;\; \TIMECONSTANT_{k-1}).
\end{equation}
Signal reconstruction for both the DoE and DoT wavelets follows from \eqref{eq:scale_bandpass_reconstruction}.
Both the DoT and DoE are wavelets, because they satisfy the admissibility criterion \eqref{eq:wavelet_admissibility}.
The DoG is non-causal and can ``react'' backwards in time, while the DoT and DoE will necessarily lag behind.
See \cref{fig:kernel_gallery} for an overview of time, frequency, and spectral decay for the three wavelet families.

\subsection{Spectral properties and frame geometry} \label{sec:spectral_properties}
Consider the frequency-to-bandwidth ratio (or quality factor) of a bandpass filter
\begin{equation} \label{eq:constant_quality}
    Q = \omega_{\rm peak} / \mathrm{BW} = \omega_{\rm peak} / (\omega_+ - \omega_-),
\end{equation}
where $\omega_\pm$ are the upper and lower $-$3 dB band edges.
The geometric spacing in \eqref{eq:scale_level_spread} implies that each channel is a temporal dilation of the previous by $c$, $\psi_{k+1}(t) = c^{-1}\psi_k(t/c)$.
In turn, the tiling is constant-Q and spans a frequency range with ratio $c^{K-1}$ \cite[Sec. 4]{pedersen2027frame}.
The spectral decay is \cite[Tab. 1]{pedersen2027frame}
\begin{equation} \label{eq:spectral_decay}
    \mathrm{DoG}: e^{-\sigma^2\omega^2/2} \quad
    \mathrm{DoT}: O(\omega^{-N}) \quad
    \mathrm{DoE}: O(\omega^{-1}).
\end{equation}
% \subsubsection{Channel overlap and Gram matrix} \label{sec:gram_matrix}
The Gram matrix $\boldsymbol{G} \in \mathbb{R}^{K \times K}$ has the entries
\begin{equation} \label{eq:gram_matrix}
    G_{jk} = \langle \psi_j, \psi_k\rangle = \frac{1}{2\pi}\int_{-\infty}^{\infty} \widehat{\psi}_j(\omega) \, \overline{\widehat{\psi}_k(\omega)} \, {\rm d}\omega,
\end{equation}
and the normalized Gram matrix is
\begin{equation} \label{eq:gram_normalized}
    G_{{\rm norm}, jk} = G_{jk} / \sqrt{G_{jj} \, G_{kk}},
\end{equation}
with $G_{jj} = \langle \psi_j, \psi_j \rangle = \|\psi_j\|^2$ \cite[Sec. 5]{pedersen2027frame}.
All three families form overcomplete, non-tight frames \cite{pedersen2027frame}.
For the $K+1$ channels here, the energy capture function \eqref{eq:energy_capture_def} instantiates as $S(\omega) = |\widehat{h}(\omega;\; \DEV_K)|^2 + \sum_{k=1}^{K} |\widehat{\psi}_k(\omega;\; \DEV_k, c)|^2$, where the lowpass term keeps $A$ from vanishing at $\omega = 0$, and the bounds are taken over the tiled band $\omega \in [0,\; 1/\DEV_0)$.
The frame bound ratio $B/A$ governs the sensitivity to noise and quantization, with the values \cite[Tab. 1]{pedersen2027frame}
\begin{equation}
    \mathrm{DoG}: 8^* \quad
    \mathrm{DoT}: 2.38 \quad
    \mathrm{DoE}: 11.63,
\end{equation}
at $c = \sqrt{2}$ and where $^*$ is a numerically evaluated quantity since the DoG bounds have no closed form.

\section{Spiking scale-covariant wavelets} \label{sec:spiking_scale-covariant_wavelets}
We define spiking wavelets by quantizing bandpass wavelet responses as streams of signed spikes, whose dual can be approximated to recover the original signal up to a bounded error that we find to be linear in the spiking threshold.

\subsection{Spiking polarity channels}
We split each channel into two polarity channels to capture sign information.
Let $\TIMECONSTANT_m$ denote the membrane time constant, distinct from the channel scale $\TIMECONSTANT_k$ since the scale selectivity resides in $\psi_k$ (we choose $\TIMECONSTANT_m = {\rm d}t$ in our simulations).
The integrative terms based on the leaky integrator \eqref{eq:li_dot} become
\begin{equation} \label{eq:spiking_channels}
    \TIMECONSTANT_m \dot{u}_k^+ = -u^+_k(t) + \bar{x}_k(t), \quad \TIMECONSTANT_m \dot{u}_k^- = -u^-_k(t) - \bar{x}_k(t),
\end{equation}
and likewise for $\bar{x}_{\rm lowpass}$ on the coarsest channel.
When $u_k^{\pm}$ crosses $\theta_{\rm thr}$, the neuron emits a spike and resets \eqref{eq:lif_srm}, acting as a quantizer with two spike trains \eqref{eq:spike_train} $z_k^+(t)$ and $z_k^-(t)$ for the positive and negative polarity channels, respectively.

\subsection{Reconstructing from polarity spikes} \label{sec:reconstruction_from_polarity_spikes}
\begin{figure*}
    \centering
    \includegraphics[width=\textwidth]{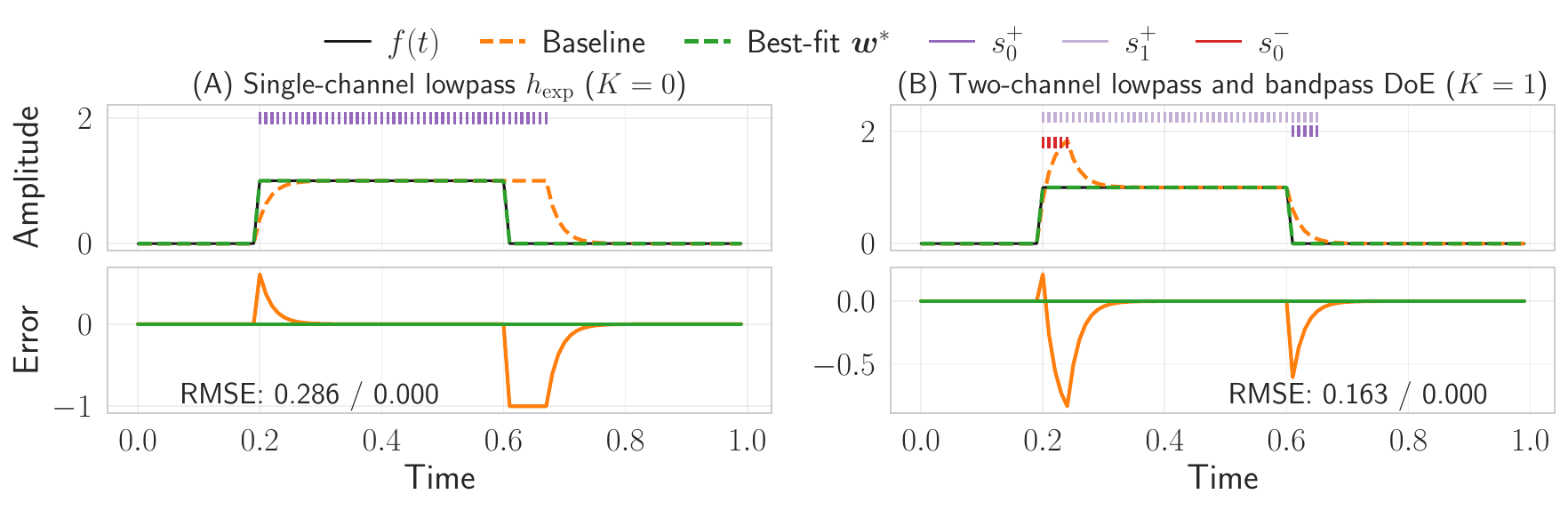}
    \caption{Reconstruction using identity weights \eqref{eq:reconstruction_weights_default} (baseline) or best-fit \eqref{eq:per_channel_decoupled_lstsq} of a boxcar signal using a lowpass filter ($K=0$, a) and a lowpass combined with a single bandpass channel DoE wavelet ($K=1$, b). RMSE is quoted per panel as baseline / best-fit.
        We use ${\rm d}t = 0.01$, $\TIMECONSTANT = 2{\rm d}t = 0.02$ and $\theta_{\rm thr} = 0.01$.
    }
    \label{fig:error_best-fit_default_box}
\end{figure*}
To reconstruct the signal, we first reconstruct each channel independently, then sum the channels to obtain the full reconstruction via the bandpass decomposition \eqref{eq:scale_bandpass_reconstruction}.
Each bandpass channel with scale $\DEV_k$ is encoded as the signed spike difference $z_k(t) = z_k^+(t) - z_k^-(t)$, which is passed through a leaky integrator $h_{\rm exp}(t; \TIMECONSTANT_r)$ \eqref{eq:truncated_exponential} with reconstruction time constant $\TIMECONSTANT_r = \TIMECONSTANT_k$.
The per-channel reconstruction kernel becomes
\begin{equation} \label{eq:channel_reconstruction}
    R_k(t;\; \DEV_k, c) = h_{\rm exp}(t;\; \TIMECONSTANT_r) * \psi(t;\; \DEV_k, c),
\end{equation}
where $\psi$ is either $\psi_{\rm DoG}$ \eqref{eq:dog_wavelet}, $\psi_{\rm DoT}$ \eqref{eq:dot_wavelet} or $\psi_{\rm DoE}$ \eqref{eq:doe_wavelet}.
This yields the following reconstruction, where $h$ is either the Gaussian \eqref{eq:gaussian}, $h_\Psi$ \eqref{eq:time-causal_limit_kernel_convolution}, or $h_{\rm exp}$ \eqref{eq:truncated_exponential}, and $z_{\rm lowpass}$ is the spike train of the low-pass filter \eqref{eq:scale-space} at scale $K$:
\begin{equation} \label{eq:spiking_wavelet_reconstruction}
    \begin{multlined}
        \tilde{f}(t) = \left(h_{\rm exp}(t;\; \TIMECONSTANT_r) * h(t;\; \DEV_K,c)\right) * z_{\rm lowpass}(t) \\
        - \sum_{k=1}^{K} R_k(t;\; \DEV_k, c) * z_k(t).
    \end{multlined}
\end{equation}
In Equation \eqref{eq:spiking_wavelet_reconstruction}, each spike contributes a uniform amplitude from the threshold $\theta_{\rm thr}$ \eqref{eq:spike_train}.

To restore the amplitude and absorb any discretization corrections, we solve for a per-spike weight by least squares.
For each channel, we build a matrix whose columns are the channel's own bandpass reconstruction kernel $R_k = h_{\rm exp}(\TIMECONSTANT_r) * \psi_k$ from \eqref{eq:channel_reconstruction}, shifted to each of channel $k$'s spike times $\mathcal{S}_k$,
\begin{equation} \label{eq:best_fit_A_spike}
    A^{(k)}_{t,i} = p_i \, R_k\!\big(t - t_{\rm spike}^{(i)};\; \sigma_k, c\big), \qquad i \in \mathcal{S}_k,
\end{equation}
where $p_i \in \{+1, -1\}$ is the polarity of spike $i$, and solve
\begin{equation} \label{eq:per_channel_decoupled_lstsq}
    \begin{split}
        \boldsymbol{w}^{(k)*} & = \underset{\boldsymbol w}{\arg \min}\big\lVert x_k - \boldsymbol{A}^{(k)}\boldsymbol w \big\rVert_2^2 \;=\; \boldsymbol{A}^{(k)\dagger} x_k, \\
        \widetilde x_k        & = \boldsymbol{A}^{(k)} \boldsymbol{w}^{(k)*}.
    \end{split}
\end{equation}
The solution is regularized.
Defining $\boldsymbol{G}^{(k)} = \boldsymbol{A}^{(k)\top}\boldsymbol{A}^{(k)}$ and $\bar{G}^{(k)}$ as the mean of its diagonal, we solve the normal equations
\begin{align} \label{eq:per_channel_decoupled_ridge}
    \boldsymbol{w}^{(k)*} & = \big(\boldsymbol{G}^{(k)} + \boldsymbol{\Gamma}^{(k)}\big)^{-1} \boldsymbol{A}^{(k)\top} x_k \nonumber \\
    \Gamma^{(k)}_{ii}     & = \rho \, \max \big(G^{(k)}_{ii}, \gamma \, \bar{G}^{(k)}\big),
\end{align}
where $\rho = 10^{-4}$ and $\gamma = 10^{-1}$ are constants.
The full reconstruction is the sum from \eqref{eq:scale_bandpass_reconstruction},
\begin{equation} \label{eq:per_spike_decoupled_reconstruction}
    \widetilde f(t) = \widetilde x_{\rm lowpass}(t) - \sum_{k=1}^{K} \widetilde x_k(t).
\end{equation}
\Cref{fig:error_best-fit_default_box} compares the identity weights \eqref{eq:reconstruction_weights_default} with the best-fit \eqref{eq:per_channel_decoupled_lstsq} on a boxcar signal at coarse time resolution ($\TIMECONSTANT = 2{\rm d}t$): a single lowpass channel ($K=0$) cannot track the boxcar edges under either weighting, while adding a bandpass channel ($K=1$) lets the spikes track the transitions and reduces both errors substantially.

This approximate reconstruction offers three advantages:\\
(1) The encoding is entirely time-recursive and no external history or state is required to propagate the filter states, so encoding runs in real time with bounded memory.\\
(2) The implementation is time-causal and implementable as differences of leaky integrators, which maps directly to neuromorphic hardware.\\
(3) The overall scheme preserves scale covariance, since both the encoding wavelets and the reconstruction filters scale covariantly with temporal scaling \eqref{eq:srm_temporal_covariance}.

\Cref{alg:encoderdecoder} lists the exact instructions to carry out the encoding and decoding proposed in this paper.
The encoding is split into a scale-space encoding and spike generation part, where the channels ($K$ bandpass channels and 1 lowpass), formed by either DoG \eqref{eq:dog_wavelet}, DoT \eqref{eq:dot_wavelet}, or DoE \eqref{eq:doe_wavelet} representations, are quantized into polarized spike trains \eqref{eq:spiking_channels} using LIF \eqref{eq:lif_srm}.
Rather than producing a bandpass representation directly from the difference kernels $\psi$, we build the bandpass representations from differences of lowpass representations \eqref{eq:difference-of-reconstruction}, which we reuse for the different bandpass scales for efficiency.
Note that the separate spike train for the lowpass channel corresponds to the signal representation $L(t;\; \DEV_K,c)$ \eqref{eq:scale-space} at the coarsest scale $K$.
The decoder first recovers the channel representations from the spike trains, from which $f$ is built by summation \eqref{eq:per_spike_decoupled_reconstruction}.
Again, the lowpass channel is processed separately from the bandpass channels to carry the coarsest signal structure.
\begin{algorithm}[h]
    \textbf{Input: } timeseries $f(t)$, wavelet kernel $h \in \{h_{\rm Gauss}, h_\Psi, h_{\rm exp}\}$, $K$, spike threshold $\theta_{\rm thr}$ \\
    \textbf{Output: } spike times $\{\mathcal{S}_1^\pm, \cdots, \mathcal{S}_K^\pm, \mathcal{S}_{\rm lowpass}^\pm\}$ (encoder), timeseries $\widetilde{f}(t)$ (decoder)
    \caption{Spiking wavelet encoding and reconstruction with lowpass and bandpass channels.}
    \begin{algorithmic}[1]
        \Statex
        \State \textbf{Scale-space lowpass encoding} \hfill \textsc{(encoder)}
        \For{$k = 0, 1, \dots, K$}
        \State $L_k(t;\sigma_k,c) \gets h(t; \sigma_k) * f(t)$ \eqref{eq:scale-space}
        \EndFor
        \Statex
        \State \textbf{Bandpass and spike generation} \hfill \textsc{(encoder)}
        \For{$k=1, \dots, K$}
        \State $\Delta L_k(t;\sigma_k, c) \gets L_k(t;\sigma_k, c) - L_k(t;\sigma_{k-1},c)$ \eqref{eq:difference-of-reconstruction}
        \State $x_k \gets \Delta L_k / \lVert \psi(t;\DEV_k,c) \rVert_2$ \eqref{eq:scale_channel_norm}
        \State $\mathcal{S}_k^\pm \gets \{t \colon u_k^\pm(t) = \theta_\mathrm{thr}\}$ \eqref{eq:spiking_channels}
        \EndFor
        \State $x_{\rm lowpass} \gets L_K / \lVert h(t;\DEV_K) \rVert_2$ \eqref{eq:scale_channel_norm}
        \State $\mathcal{S}^{\pm}_{K,\rm lowpass} \gets \{t \colon u_{\rm lowpass}^{\pm}(t) = \theta_{\rm thr}\}$ \eqref{eq:spiking_channels}
        \LineComment{LIF on lowpass channel $L(t;\DEV_K,c)$}
        \Statex
        \State \textbf{Per-channel decoding}  \hfill \textsc{(decoder)}
        \For{$k = 1, \dots, K$}
        \State $A^{(k)}_{t, i} \gets p_i R_k\left(t - t_{\rm spike}^{(i)}\right),\ i \in \mathcal{S}^{\pm}_k$ \eqref{eq:best_fit_A_spike}
        \State $\boldsymbol{w}^{(k)*} \gets \arg\min_{\boldsymbol w} \left \lVert \Delta L_k(t;\DEV_k,c) - \boldsymbol{A}^{(k)}\boldsymbol w \right\rVert_2^2$ \eqref{eq:per_channel_decoupled_lstsq}
        \State $\widetilde{\Delta L}_k \gets \boldsymbol{A}^{(k)} \boldsymbol{w}^{(k)*}$
        \EndFor
        \State $A^{(\rm lp)}_{t, i} \gets p_i R_{K,\rm lowpass}\left(t - t_{\rm spike}^{(i)}\right),\ i \in \mathcal{S}^{\pm}_{K,\rm lowpass}$
        \State $\boldsymbol{w}^{(\rm lp)*} \gets \arg\min_{\boldsymbol w} \left \lVert L_{K,\rm lowpass}(t;\DEV_K,c) - \boldsymbol{A}^{(\rm lp)}\boldsymbol w \right\rVert_2^2$
        \State $\widetilde{L}_{K,\rm lowpass}(t;\DEV_K,c) \gets \boldsymbol{A}^{(\rm lp)} \boldsymbol{w}^{(\rm lp)*}$
        \LineComment{Low-pass spike channel decoding}
        \Statex
        \State \textbf{Signal synthesis} \hfill \textsc{(decoder)}
        \State  $\widetilde{f}(t) \gets \widetilde{L}_{K,\rm lowpass}(t;\DEV_K,c) - \sum_{k=1}^K \widetilde{\Delta L}_k(t; \DEV_k, c)$ \eqref{eq:per_spike_decoupled_reconstruction}
    \end{algorithmic}
    \label{alg:encoderdecoder}
\end{algorithm}

\subsection{Reconstruction and quantization errors} \label{sec:reconstruction_error}
% The bandpass decomposition enables signal reconstruction via \eqref{eq:scale_bandpass_reconstruction}, where the original signal is recovered from the lowpass residual along with the sum of the bandpass channels.
The quality of the reconstruction depends on the spectral decay of the bandpass channels and the quantization error introduced by the spiking encoding.
The faster spectral decay for the DoT \eqref{eq:spectral_decay} means that each DoT channel has a well-defined effective bandwidth $\Omega_k$, which guarantees that the spectral mass outside $\Omega_k$ is bounded for any desired tolerance.
For the DoE, the slower decay requires additional assumptions on the signal.

There are two sources of error in the spiking case: a mismatch in amplitude introduced by the discrete spikes and a mismatch in time, introduced by the simulated temporal discretization.
We begin by studying the continuous case, where the spectral decay of time-scaled responses provides bandwidth limits for each channel which, in turn, bounds the amplitude error for the individual channels.
We then proceed to study the temporal discretization, which we combine with the amplitude error above to bound the overall spiking reconstruction process.

\subsubsection{Spike amplitude error}
Let $ \Omega$ be the \emph{bandwidth} of signal $f$, and $ \neg \Omega = \{\omega\colon |\omega| > \Omega\}$ be the region outside it, not restricted to the $-$3 dB margin in \eqref{eq:constant_quality}.
The spectral mass in $ \neg \Omega$ is bounded from above by some tolerance $\Delta_\Omega$ \cite[Sec.~2.4]{carbajal2026model}.
For each bandpass channel, the filtered signal $\widehat{\Delta L}_k(\omega;\; \DEV_k, c)$ satisfies
\begin{align} \label{eq:bandwidth_tail_integral}
    \int_{\omega \in \neg\Omega_k} \!\!\! \!\!\!\!\!\!\!\!\! |\widehat{\Delta L}_k(\omega;\; \DEV_k, c)| \, {\rm d}\omega 
    & = \int_{\omega \in \neg\Omega_k} \!\!\! \!\!\!\!\!\!\!\!\!|\widehat{\psi}(\omega;\; \DEV_k, c)| \, |\widehat{f}(\omega)| \, {\rm d}\omega \nonumber \\
    & \leqslant C \int_{\omega \in \neg\Omega_k} \!\!\! \!\!\!\!\!\!\!\!\! | \omega |^{-N}\, |\widehat{f}(\omega)| \, {\rm d} \omega,
\end{align}
for some $C \in \mathbb{R}^+$, where the last step holds for $|\omega|$ sufficiently large by \eqref{eq:spectral_decay}.
A finite $\Omega_k$ meeting the tolerance $\Delta_\Omega$ exists whenever the right-hand side is finite and decreasing in $\Omega_k$.
That depends on $N$ and on the decay of $\widehat{f}$ jointly, which separates the three families.

For $N = 1$ (DoE), the $O(|\omega|^{-1})$ decay of the bandpass kernel is not sufficient to guarantee convergence
of~\eqref{eq:bandwidth_tail_integral} from the kernel alone.
Additional assumptions on $\widehat{f}$, or the use of reconstruction error feedback mechanisms, are required to ensure that the approximate bandwidth condition is met.
For the non-causal DoG, the Gaussian decay \eqref{eq:spectral_decay} dominates, so the tail integral \eqref{eq:bandwidth_tail_integral} converges to an essential bandwidth.

Assume the bandpass signal $\Delta L_k(t;\; \DEV_k, c)$ has essential bandwidth $\Omega_k$ at tolerance $\Delta_\Omega = \theta_{\rm thr}$, where $\theta_{\rm thr}$ is the firing threshold in \eqref{eq:spiking_channels}.
A spectral tail carrying less than one threshold of charge is invisible to the encoder.
Any two signals producing the same spike train have running integrals (e.g.\ the leaky integrator \eqref{eq:li_dot}) agreeing to within $\theta_\mathrm{thr}$ and differentiation of a signal with bandwidth $\Omega_k$ increases its supremum by at most a factor $\Omega_k$.
This bounds the per-channel amplitude encoding error \cite[Th. 2.2]{carbajal2026model}
\begin{equation} \label{eq:channel_error_amp}
    \lVert \epsilon_{{\rm amp}, k} \rVert_\infty =
    \lVert \Delta L_k(t;\; \DEV_k, c) - \widetilde{\Delta L}_k(t;\; \DEV_k, c)\rVert_\infty \leqslant C \, \theta_{\rm thr} \, \Omega_k,
\end{equation}
where $C > 0$ is a constant absorbing leakage and the tail outside the band.
This bound is model-agnostic and holds for any (spiking) reconstruction method.

The error $\epsilon$ between the original signal $f$ and the reconstruction $\tilde{f}$ from \eqref{eq:scale_bandpass_reconstruction} is then
\begin{equation} \label{eq:total_reconstruction_error}
    \begin{aligned} f(t) - \widetilde{f}(t) & = L_K - \widetilde{L}_K - \sum_{k=1}^K \left( \Delta L_k - \widetilde{\Delta L}_k\right) \\
    & = \epsilon_{{\rm amp}, K} - \sum_{k=1}^K \epsilon_{{\rm amp}, k}.\end{aligned}
\end{equation}
Inserting \eqref{eq:channel_error_amp}, where the lowpass residual contributes the essential bandwidth $Q \, \Omega_K$ \eqref{eq:constant_quality} with the constant-$Q$ factor absorbed into $C$, we get the total reconstruction error
\begin{equation}  \label{eq:total_error_amp}
    \lVert f(t) - \widetilde{f}(t)\rVert_\infty \leqslant C\ \theta_{\rm thr}\ \left( \Omega_K + \sum_{k=1}^K \Omega_k \right).
\end{equation}
Given that the bandpass filters are self-similar over scales, a reasonable measure of the bandwidth $\Omega_k$ of the bandpass filters must obey the following variability over scales
\begin{equation}
    \Omega_k = \Omega_1 / c^{k-1},
\end{equation}
which when inserted into \eqref{eq:total_error_amp} by the sum of a geometric series gives the closed-form expression
\begin{equation} \label{eq:total_error_bound}
    \lVert f(t) - \widetilde{f}(t)\rVert_\infty \leqslant C\ \theta_{\rm thr}\ \left( \frac{\Omega_1}{c^{K-1}} + \Omega_1 \left(  \frac{1 - \frac{1}{c^{K}}}{1 - \frac{1}{c}} \right) \right).
\end{equation}
For $K \rightarrow \infty$ this gives the error bound
\begin{equation}
    \label{eq:limit-error-bound}
    \lVert f(t) - \widetilde{f}(t)\rVert_\infty \leqslant C\ \theta_{\rm thr} \left( \frac{\Omega_1 \, c}{c - 1}\right).
\end{equation}
The bound decreases with $c$ because it counts quantization error but not rate, so it cannot select $c$ or $K$.
That trade-off is measured in \cref{sec:rate_distortion}.

\subsubsection{Discretization error}
In simulations, we are limited to a time-granularity of $\Delta t$.
Additionally, the temporal discretization introduces another error from the temporal uncertainty of the spike times: in a causal simulation, a spike firing at $t_0 + \epsilon$, for some number $\epsilon > 0$, is only registered at $t_0 + \Delta t$, giving a worst-case delay of $\Delta t$ per spike.
For $m_k$ spikes in a channel $k$, the per-channel error, combining amplitude quantization \eqref{eq:channel_error_amp} and temporal discretization is
\begin{equation} \label{eq:channel_error_disc}
    \|\epsilon_{{\rm disc}, k}\|_\infty \!= \|\epsilon_{{\rm amp}, k}\|_\infty\! +\! m_k\ \Delta t\ \Omega_k \leqslant C \theta_{\rm thr} \, \Omega_k + C \, m_k \, \Delta t \, \Omega_k.
\end{equation}
The total reconstruction from a time-discretized reconstruction $\widetilde{f}_{\rm disc}$ is then
\begin{align} \label{eq:channel_error_amp_and_disc}
     & \|f(t) - \widetilde{f}_{\rm disc}(t)\|_\infty \leqslant \|\epsilon_{{\rm amp}, K}\|_\infty + \sum_{k=1}^K \|\epsilon_{{\rm disc}, k}\|_\infty \nonumber \\
     & \leqslant C\ \theta_{\rm thr}\ \left(\Omega_K + \sum_{k=1}^K \Omega_k\right) + C \, \Delta t \sum_{k=1}^K m_k \, \Omega_k.
\end{align}

\section{Experiments} \label{sec:experiments}

We test the bandpass and spiking wavelets against three properties that an encoder-decoder pair should satisfy: frequency localization, scale covariance, and formal bounds for stable reconstruction.
Frequency localization follows analytically from the channel tiling and channel overlap (see \cref{sec:spectral_properties}), so the experiments below address scale covariance and the reconstruction bounds.
The reconstruction is measured against two baseline continuous-time wavelets and two event encoders.
For the continuous-time wavelets, we use the non-causal Morlet wavelet that scales and shifts Gaussians \cite[(3.3.26)]{daubechies1992ten}, and the causal Szu wavelet that are restricted to observing past inputs \cite[(10)]{szu1992causal}.

For the event encoders we use $\Sigma\Delta$ \eqref{eq:sigma_delta} and a multi-channel integrate-and-fire time-encoding machine (MC-IFTEM) \cite{adam2020sampling}.
MC-IFTEM fires the $m$th of its $M$ integrate-and-fire channels at trigger times $t^{m}_k$ set by the t-transform that maps amplitudes into the time domain~\cite[(2)]{adam2020sampling}
\begin{equation} \label{eq:mc_iftem}
    \int_{t^{m}_{k}}^{t^{m}_{k+1}} \big( f(t) + \beta \big) \, {\rm d}t = 2\kappa_{\rm tem} \, \theta_{\rm tem},
    \qquad m = 1,\dots,M,
\end{equation}
where $\beta$ is a bias, $\kappa_{\rm tem}$ the integrator constant, and $\theta_{\rm tem}$ the threshold ($\delta$ in \cite{adam2020sampling}).
% The bias must exceed the signal bound, $\beta > \lVert f \rVert_\infty$, which keeps the integrand positive and bounds the trigger separation above by $2\kappa_{\rm tem}\theta_{\rm tem} / (\beta - \lVert f \rVert_\infty)$ \cite[(1),(6)]{adam2020sampling}.
% The identical $M$ channels start integration at equally spaced offsets in $[0, 2\theta_{\rm tem})$, which interleaves their trigger times.
MC-IFTEM recovers $f(t)$ from the $M$ trains by solving a dense frame system for the coefficients $\boldsymbol{a}$ of the interval-integral sinc frame against the measurements $\boldsymbol{q}$,
\begin{equation} \label{eq:iftem_decode}
    G \, \boldsymbol{a} = \boldsymbol{q}, \qquad G \in \mathbb{R}^{P \times P}, \quad P = |\mathcal{S}| - M
\end{equation}
one measurement per interspike interval, at $\mathcal{O}(|\mathcal{S}|^3)$ cost in the pooled event-trigger count $|\mathcal{S}|$.

A $\Sigma\Delta$ modulator \cite{carbajal2026model} instead integrates the difference between $f[t]$ and its one-bit feedback $b[t]$ through two states $v_1, v_2$, binarized by $\operatorname{sign}$ (second-order form) \cite[(11)]{aziz1996overview}
\begin{align} \label{eq:sigma_delta}
    v_1[t] & = v_1[t-1] + f[t] - b[t-1], \nonumber   \\
    v_2[t] & = v_2[t-1] + v_1[t] - b[t-1], \nonumber \\
    b[t]   & = \operatorname{sign}\big(v_2[t]\big).
\end{align}
The event encoders are compared against the wavelets at a matched spike rate.
For an encoder $E$ and a signal $f(t)$, let $\mathcal{S}_E(f)$ be the set of events over channels ($2(K+1)$ for the polarized spiking wavelet \eqref{eq:spiking_channels}, the $M$ trigger trains in the MC-IFTEM \eqref{eq:mc_iftem}, or the one-bit output of $\Sigma\Delta$ \eqref{eq:sigma_delta}).
The spike rate is the event count per input sample, averaged over the evaluation set $\mathcal{D}$,
\begin{equation} \label{eq:spike_rate}
    \bar{r}(E) = \frac{1}{|\mathcal{D}|} \sum_{f \in \mathcal{D}}
    \frac{\big|\mathcal{S}_E(f(t))\big|}{\lvert f \rvert},
\end{equation}
where $\lvert f \rvert$ is the number of samples in $f(t)$, i.e. one second at the dataset's sampling rate ($\lvert f \rvert = 360$ for MIT-BIH, $\lvert f \rvert = 16\,000$ for LibriSpeech).
The Nyquist-rate sampling of a signal band-limited to $f_{\max}$
acquires $2f_{\max}$ samples per second \cite[Thm.~1]{shannon1949communication},
i.e., $2f_{\max} \, \Delta t$ events per timestep in the units of \eqref{eq:spike_rate}.
In our experiments, $f_{\max} = 1/(2\Delta t)$ is the highest frequency representable at each dataset's sampling rate (180~Hz for MIT-BIH, 8~kHz for LibriSpeech), so the Nyquist rate is exactly one event per timestep.
Each baseline is a one-parameter family $E_\lambda$, with $\lambda$ the oversampling ratio for $\Sigma\Delta$ and the threshold $\theta_{\rm tem}$ \eqref{eq:mc_iftem} for MC-IFTEM, and $\bar{r}(E_\lambda)$ is monotone in $\lambda$.
Spike-rate matching selects $\lambda^\star$ such that
\begin{equation} \label{eq:rate_matching}
    \big|\bar{r}(E_{\lambda^\star}) - \bar{r}(E_\mathrm{ref}) \big| \leqslant \varepsilon \, \bar{r}(E_\mathrm{ref}), \qquad \varepsilon = 0.05,
\end{equation}
by bisection on $\lambda$, with $E_\mathrm{ref}$ the spiking DoT at $\theta_\mathrm{thr} = 0.1$.
This keeps the number of events per input sample on-par with the DoT reference wavelet, comparing fidelity at a controlled event rate.

% \begin{figure}
%     \centering
%     \includegraphics[width=\linewidth]{figures/example_signals.pdf}
%     \caption{Sample signals from the MIT-BIH ECG and LibriSpeech datasets.
%         The signals have been standardized and clipped to 1 second duration.}
%     \label{fig:sample_signals}
% \end{figure}
The experiments are conducted on the MIT-BIH heart arrhythmia dataset \cite{moody2001impact} and the LibriSpeech audiobook recordings \cite{panayotov2015librispeech}.
The datasets are chosen for their different characteristics (see \cref{fig:realistic_encoding} for signal samples): MIT-BIH ECG features bursts of activity sampled at 360~Hz, while LibriSpeech contains dense speech signals sampled at 16~kHz.
We split the data into one-second samples.

\subsection{Experimental details} \label{sec:experimental_details}

All simulations are clocked to the native sampling rate of each dataset.
The DoG \eqref{eq:dog_wavelet}, DoE \eqref{eq:doe_wavelet}, and DoT \eqref{eq:dot_wavelet} wavelets represent the signal exactly.
The continuous wavelet transforms (CWT) and spiking kernels are continuous in time, but evaluated on a discrete grid, introducing discretization error that vanishes at $\Delta t \to 0$.
We choose $\Delta t$ as the inverse of the sampling rate, $\Delta t = \{1/360, 1/16000\}$ for MIT-BIH and LibriSpeech, respectively.
In all experiments, the DoT wavelet uses a finite cascade of $N = 7$ truncated-exponential stages \eqref{eq:time-causal_limit_kernel_convolution} to approximate the time-causal limit kernel.
For fixed bandwidth minimum and maximum $[\Omega_{\min}, \Omega_{\max}]$, we set the scale ratio $c$ according to \cite[(29)]{pedersen2027frame}.

As error metric, we rely on the root mean squared error (RMSE) standardized to zero mean and unit variance.
Standardization permits us to compare reconstruction error across LibriSpeech and MIT-BIH ECG datasets.
For non-spiking wavelets, $\widetilde{f}$ comes from frame synthesis \eqref{eq:frame_reconstruction}.
For spiking systems, $\widetilde{f}$ is the per-spike least-squares reconstruction \eqref{eq:best_fit_A_spike}.

The probe of the scale-covariance experiment \eqref{eq:gabor_probe} uses $n_{\rm cyc} = 3$ and is centered at one tenth of the window.
We place the carrier at $f_0 = 72$~Hz for MIT-BIH and $f_0 = 3$~kHz for LibriSpeech, high in each band so that dilation has room to move the activity toward the coarser channels.

\begin{figure*}[t]
    \centering
    \includegraphics[width=\textwidth]{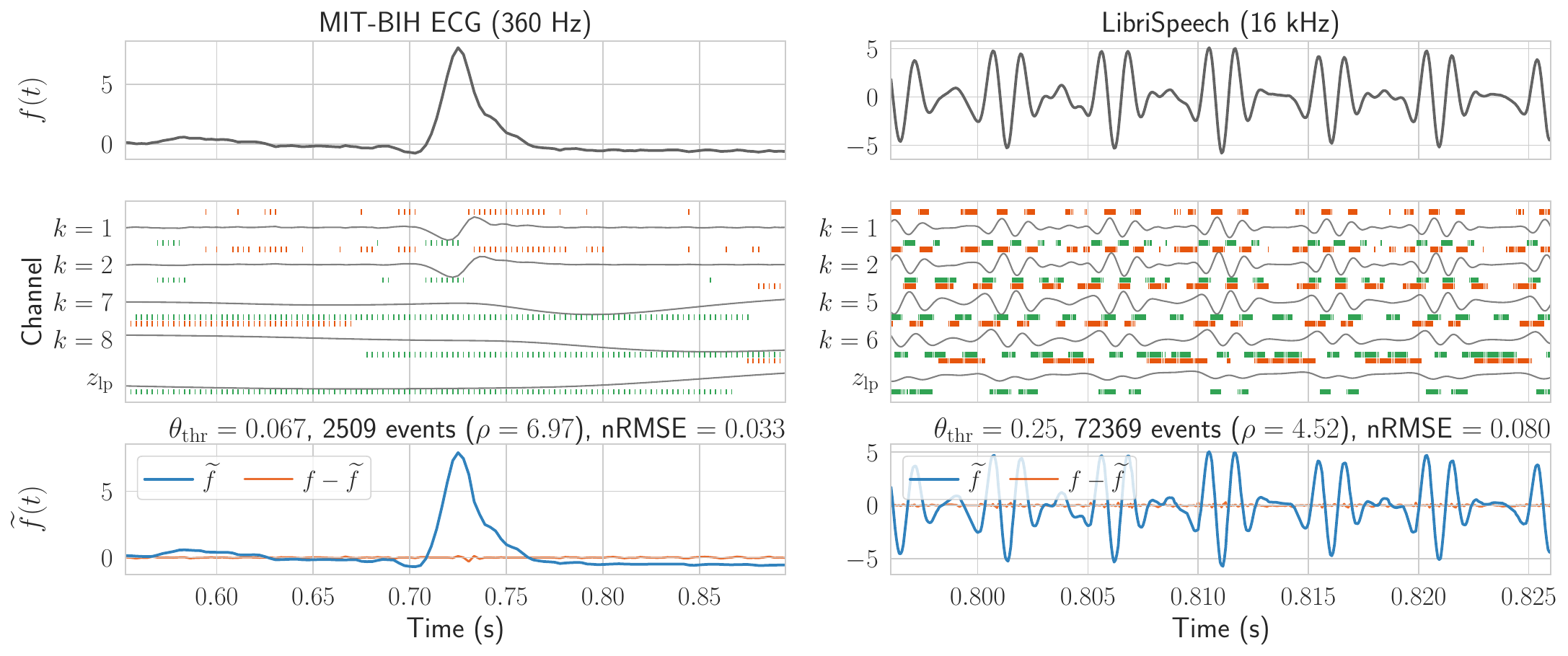}
    \caption{Spiking DoT encoding and reconstruction of real signals, at the rates \cref{tab:reconstruction} reports for $c = 2$.
        \textbf{Top}: the signal $f(t)$.
        \textbf{Middle}: polarity spikes $z_k^\pm$ \eqref{eq:spike_train} for the two finest and two coarsest bandpass channels and the lowpass residual, with the normalized channel response $\bar{x}_k$ \eqref{eq:scale_channel_norm} drawn behind each row, scaled individually. The threshold crossing itself is shown in \cref{fig:error_best-fit_default_box}.
        \textbf{Bottom}: the reconstruction $\widetilde{f}$ \eqref{eq:per_spike_decoupled_reconstruction} and the residual $f - \widetilde f$.
    }
    \label{fig:realistic_encoding}
\end{figure*}

We chose the MIT-BIH ECG \cite{moody2001impact} and LibriSpeech \cite{panayotov2015librispeech} datasets because they express two distinct characteristics.
The MIT-BIH dataset consists of heart arrhythmia recordings that mostly oscillate around 0, with short, transient bursts.
The LibriSpeech contains recordings of audiobooks by English speakers.
\Cref{fig:realistic_encoding} visualizes the different nature of the signals.
The LibriSpeech has much more structure and requires a denser sampling to recover the 16 kHz signal.

The experimental details and code are openly available at \href{https://github.com/jegp/swavelet}{github.com/jegp/swavelet}.
The spiking DoE and DoT encoders additionally export to the Neuromorphic Intermediate Representation \cite{pedersen2024neuromorphic} as a graph of leaky-integrator, linear, and LIF primitives for deployment on neuromorphic backends.

\subsection{Reconstruction experiment} \label{sec:reconstruction_experiment}
Using \cref{alg:encoderdecoder}, we measure the ability of the wavelet systems to reconstruct signals from their respective encodings.
We group the wavelets into four classes: bandpass wavelets (DoG, DoE, and DoT), continuous wavelets (CWT), spiking wavelets, and event encoders (MC-IFTEM and $\Sigma\Delta$).
The spiking DoG serves as a non-causal control: it shares the full encoding and decoding pipeline with the causal spiking DoE and DoT, so their comparison isolates the cost of causality within the same construction.
For the non-spiking DoG \eqref{eq:dog_wavelet}, DoE \eqref{eq:doe_wavelet}, and DoT \eqref{eq:dot_wavelet} wavelets, the reconstruction is exact by construction.
For the Morlet wavelets (non-causal) and the Szu wavelets (causal), we compute dense coefficients $c_k(t) = \langle x, \psi_k(\cdot - t)\rangle$ for each channel $k$ and reconstruct via a least-squares pseudoinverse.

All event encoders are decoded by the same oracle regularized least squares \eqref{eq:per_channel_decoupled_ridge} fitted against the clean signal.
The spiking wavelet fits its per-spike weights onto the bandpass channel outputs $x_k$, $\Sigma\Delta$ its moving-average atoms, and the MC-IFTEM its sinc frame.
The reconstructions therefore upper-bound what a decoder using these atoms can recover from the spike trains.
All event encoders are compared across the same event crossing threshold.
The spiking wavelets share the firing threshold $\theta_{\rm thr}=0.1$, and the MC-IFTEM threshold $\theta_{\rm tem}$ and the $\Sigma\Delta$ oversampling ratio are set once by rate-matching \eqref{eq:rate_matching} to the spiking DoT.
The event rates are therefore an outcome of the encoding, reported per entry in \cref{tab:reconstruction}.

\begin{table}
    \centering
    \caption{Reconstruction RMSE across 100 samples from the MIT-BIH ECG \cite{moody2001impact} and LibriSpeech \cite{panayotov2015librispeech} datasets.
        Event rates \eqref{eq:spike_rate} are reported in parentheses.
    }
    \label{tab:reconstruction}
    % Generated by neurips/experiment_encoding_efficiency.py (writes figures/experiment_encoding_table.tex); pasted verbatim.
    \begin{tabular*}{\columnwidth}{@{\extracolsep{\fill}}l r@{\,}l r@{\,}l@{}}
        \toprule
        & \multicolumn{4}{c}{MIT-BIH} \\
        \cmidrule(lr){2-5}
        Wavelet & \multicolumn{1}{c}{$c$=2, $K$=8} &  & \multicolumn{1}{c}{$c$=$\sqrt2$, $K$=15} &  \\
        \midrule
        DoG/DoE/DoT & \multicolumn{1}{c}{exact} &  & \multicolumn{1}{c}{exact} &  \\
        \midrule
        Morlet & $0.079\,{\scriptstyle\pm\,0.038}$ &  & $0.060\,{\scriptstyle\pm\,0.035}$ &  \\
        Szu (causal) & $0.291\,{\scriptstyle\pm\,0.096}$ &  & $0.190\,{\scriptstyle\pm\,0.095}$ &  \\
        \midrule
        Spiking DoG & $0.068\,{\scriptstyle\pm\,0.043}$ & {\scriptsize(6.2)} & $0.068\,{\scriptstyle\pm\,0.044}$ & {\scriptsize(11.3)} \\
        Spiking DoE & $0.085\,{\scriptstyle\pm\,0.042}$ & {\scriptsize(5.9)} & $0.118\,{\scriptstyle\pm\,0.051}$ & {\scriptsize(8.9)} \\
        Spiking DoT & {\bfseries\boldmath $0.059\,{\scriptstyle\pm\,0.035}$} & {\scriptsize(7.0)} & $0.070\,{\scriptstyle\pm\,0.036}$ & {\scriptsize(11.6)} \\
        \midrule
        MC-IFTEM & $0.096\,{\scriptstyle\pm\,0.168}$ & {\scriptsize(6.3)} & {\bfseries\boldmath $0.047\,{\scriptstyle\pm\,0.109}$} & {\scriptsize(11.3)} \\
        $\Sigma\Delta$ & $0.181\,{\scriptstyle\pm\,0.118}$ & {\scriptsize(7.0)} & $0.106\,{\scriptstyle\pm\,0.065}$ & {\scriptsize(12.0)} \\
        \midrule
        & \multicolumn{4}{c}{LibriSpeech} \\
        \cmidrule(lr){2-5}
        & \multicolumn{1}{c}{$c$=2, $K$=6} &  & \multicolumn{1}{c}{$c$=$\sqrt2$, $K$=12} &  \\
        \midrule
        DoG/DoE/DoT & \multicolumn{1}{c}{exact} &  & \multicolumn{1}{c}{exact} &  \\
        \midrule
        Morlet & $0.079\,{\scriptstyle\pm\,0.081}$ &  & {\bfseries\boldmath $0.064\,{\scriptstyle\pm\,0.074}$} &  \\
        Szu (causal) & $0.529\,{\scriptstyle\pm\,0.123}$ &  & $0.355\,{\scriptstyle\pm\,0.122}$ &  \\
        \midrule
        Spiking DoG & $0.072\,{\scriptstyle\pm\,0.040}$ & {\scriptsize(3.5)} & $0.070\,{\scriptstyle\pm\,0.040}$ & {\scriptsize(6.3)} \\
        Spiking DoE & $0.092\,{\scriptstyle\pm\,0.047}$ & {\scriptsize(3.5)} & $0.139\,{\scriptstyle\pm\,0.059}$ & {\scriptsize(4.8)} \\
        Spiking DoT & {\bfseries\boldmath $0.067\,{\scriptstyle\pm\,0.041}$} & {\scriptsize(4.6)} & $0.103\,{\scriptstyle\pm\,0.115}$ & {\scriptsize(7.4)} \\
        \midrule
        $\Sigma\Delta$ & $0.277\,{\scriptstyle\pm\,0.048}$ & {\scriptsize(4.0)} & $0.138\,{\scriptstyle\pm\,0.026}$ & {\scriptsize(8.0)} \\
        \bottomrule
    \end{tabular*}
\end{table}

\Cref{tab:reconstruction} shows the results for the exact bandpass wavelets (DoG, DoE, and DoT), as well as the continuous, but discretized, wavelets (Morlet, Szu, and spiking versions of DoG, DoE, and DoT) and the event-encoder baselines (MC-IFTEM~\eqref{eq:mc_iftem} and $\Sigma\Delta$~\eqref{eq:sigma_delta}).
Despite the spike quantization, the spiking wavelets stay close to the non-causal Morlet baseline, with the DoT matching or exceeding it at $c=2$.
This is notable because the spiking DoT carries both quantization and discretization error, whereas the non-causal Morlet's error is purely due to discretization from the digital sampling.

Moving from $c=2$ to the denser $c=\sqrt{2}$ grid leaves the spiking DoG essentially unchanged, while the error of the spiking causal DoE and DoT wavelets grows.
This is attributed to the increasing overlap between channels: as $c \to 1$ the bandpass responses overlap more strongly, the off-diagonal entries of the normalized Gram matrix \eqref{eq:gram_normalized} grow, and the frame bound ratio $B/A$ rises (see \cref{sec:spectral_properties}).

The causal Szu wavelet is subject to the same sampling issues as the Morlet wavelet, but the causality requirement incurs a $3$--$7\times$~RMSE penalty.
This strengthens the results of the spiking DoE and DoT wavelets, that are both causal and competitive in the reconstruction metric.

Of the event encoders in \cref{tab:reconstruction}, MC-IFTEM matches the spiking wavelets but $\Sigma\Delta$ does not, trailing the best spiking wavelet by between $1.5$ and $4\times$ at the same event rate.
For the MIT-BIH ECG, the matched rate yields ${\sim}10^3$ events, small enough that the $\mathcal{O}(|\mathcal{S}|^3)$ joint decode \eqref{eq:iftem_decode} is tractable.
For LibriSpeech it reaches $|\mathcal{S}|\sim 10^5$ events, at which the decode is intractable; we therefore omit MC-IFTEM there.

\begin{figure}
    \centering
    \includegraphics[width=\linewidth]{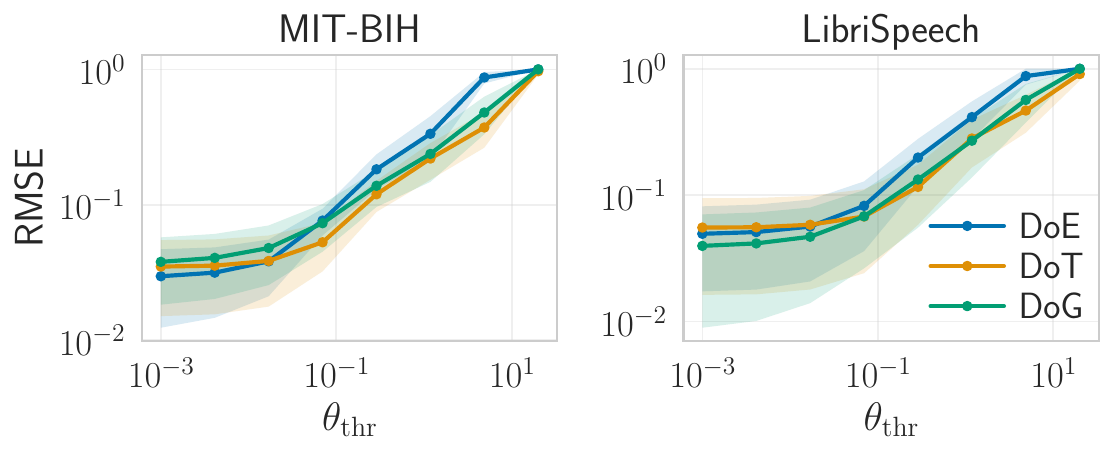}
    \caption{
        RMSE versus $\theta_{\rm thr}$, bounded linearly by $\theta_{\rm thr}$ following
        \eqref{eq:limit-error-bound}.
        The graded area shows one standard deviation of the RMSE across 10 samples.
    }
    \label{fig:threshold_sweep}
\end{figure}
\Cref{fig:threshold_sweep} sweeps the firing threshold $\theta_\mathrm{thr}$: the measured RMSE stays within the linear envelope of the reconstruction bound \eqref{eq:limit-error-bound}, demonstrating the predicted linear dependence.
Below $\theta_\mathrm{thr} \approx 0.02$ the curves flatten to a level determined by the temporal discretization \eqref{eq:channel_error_disc}.

\subsection{Spike rate} \label{sec:rate_distortion}
\begin{figure}[t]
    \centering
    \includegraphics[width=\linewidth]{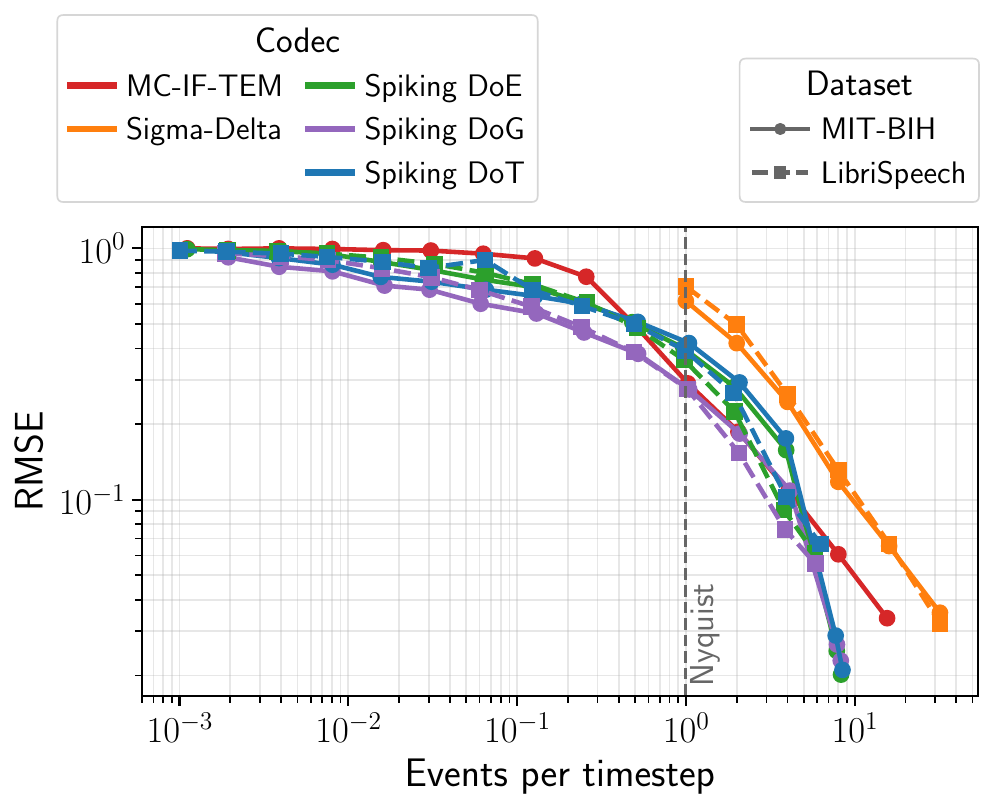}
    \caption{Operational rate-distortion: reconstruction RMSE versus events per timestep for the spiking wavelets (DoT, DoE, DoG), a second-order $\Sigma\Delta$ modulator~\eqref{eq:sigma_delta}, and the MC-IFTEM~\eqref{eq:mc_iftem}, on MIT-BIH (solid) and LibriSpeech (dashed).
        Every codec is matched onto the same grid of target rates \eqref{eq:rate_matching}.
        The dashed vertical marks the Nyquist rate $2 f_{\max}\Delta t = 1$ \eqref{eq:spike_rate}.
        MC-IFTEM is shown on MIT-BIH only, as its $\mathcal{O}(|\mathcal{S}|^3)$ joint decode~\eqref{eq:iftem_decode} is intractable at the 16kHz LibriSpeech rate.}
    \label{fig:rate_distortion}
\end{figure}

\cref{tab:reconstruction} fixes a single firing threshold, but any encoder can spend more spikes to lower its error, so one operating point is only a snapshot.
We therefore report the operational rate-distortion: reconstruction RMSE as a function of the spike rate, the code cost in events per timestep, which asks which encoder reaches a given fidelity at the lowest rate.
\cref{fig:rate_distortion} plots this trade-off for the spiking wavelets, a second-order $\Sigma\Delta$ modulator~\eqref{eq:sigma_delta} \cite{carbajal2026model}, and MC-IFTEM~\eqref{eq:mc_iftem}, all at $c=2$.
The sweep therefore does not cover the $c=\sqrt{2}$ setting at which MC-IFTEM leads \cref{tab:reconstruction}, where it encodes with $M=16$ shifted integrators rather than the $M=9$ used here.
The $\Sigma\Delta$ curve stops at one event per timestep: the modulator \eqref{eq:sigma_delta} emits one bit per input sample at its lowest rate setting, and every emitted bit counts as an event.

\begin{figure*}[t]
    \centering
    \begin{subfigure}{.49\linewidth}
        \includegraphics[width=\linewidth]{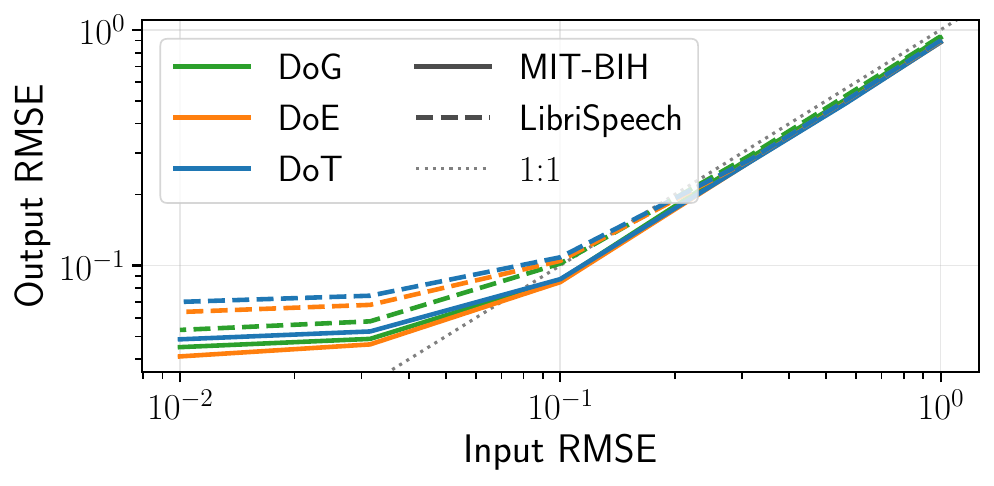}
    \end{subfigure}\hfill
    \begin{subfigure}{.49\linewidth}
        \includegraphics[width=\linewidth]{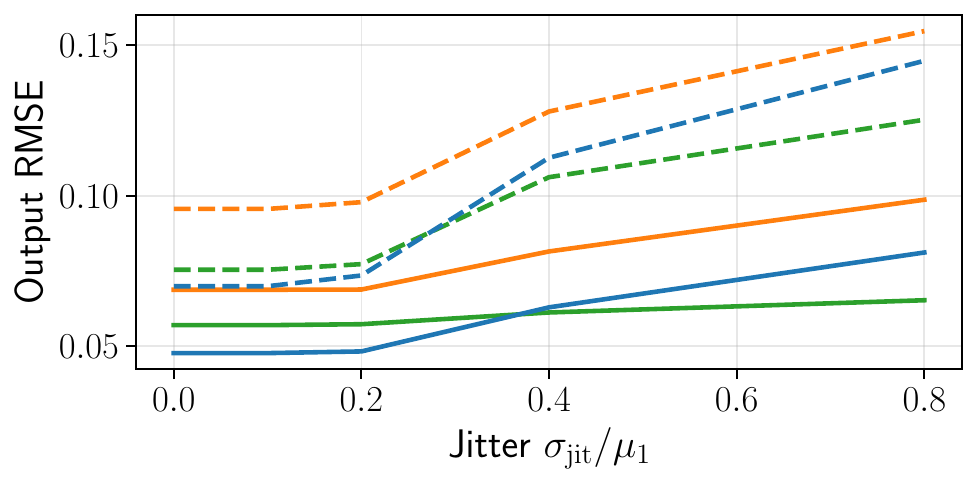}
    \end{subfigure}
    \caption{Robustness of the spiking wavelets, scored against the clean reference.
        \textbf{Left}: output RMSE versus input RMSE under additive white Gaussian noise.
        \textbf{Right}: output RMSE versus Gaussian spike-timing jitter of standard deviation
        $\sigma_{\rm jit}$, as a fraction of the finest channel scale $\mu_1$.}
    \label{fig:robustness}
\end{figure*}

The spiking wavelets improve steadily with rate throughout.
They reach a given error at one and a half to four times fewer events than $\Sigma\Delta$, which cannot encode below the Nyquist rate at all, while MC-IFTEM matches them below ${\sim}4$ events per timestep on MIT-BIH and falls behind above it.

\subsection{Robustness} \label{sec:robustness}
The reconstruction bound \eqref{eq:total_error_bound} holds for any finite-energy input, so the encoding should degrade gracefully under noise.
To test this, we add white Gaussian noise to the input across a range of noise levels, shown as input RMSE, and score the reconstruction $\widetilde{f}$ in RMSE against the clean reference (\cref{fig:robustness}, left).
Both axes are amplitude ratios in RMSE, so an identity codec would sit on the $1{:}1$ line.
The output RMSE tracks the input RMSE roughly one-to-one down to ${\approx}0.1$, then saturates at the codec's own reconstruction floor, $0.049$ on ECG and $0.070$ on speech.
The encoding therefore neither amplifies the input noise nor saturates above that floor, which is a precondition for use on real, noisy signals.

Especially on hardware, spike times are subject to jitter, which can be modeled as a random perturbation of the spike times.
We perturb the encoded spike-times with a zero-mean Gaussian jitter of standard deviation $\sigma_{\rm jit}$, expressed as a fraction of the finest channel scale $\mu_1$, shown in \cref{fig:robustness}, right.
The output RMSE stays constant while $\sigma_{\rm jit}$ stays under $\sim 0.1$ of the finest channel scale $\TIMECONSTANT_1$, then grows sub-linearly in $\sigma_{\rm jit}$, staying within about twice the jitter-free error throughout.
A spike displaced by $\sigma_{\rm jit}$ contributes a per-channel error of order $\sigma_{\rm jit} / \mu_k$, so the coarse channels are intrinsically robust and the finest channel dominates.

\begin{figure}[t]
    \centering
    \includegraphics[width=\linewidth]{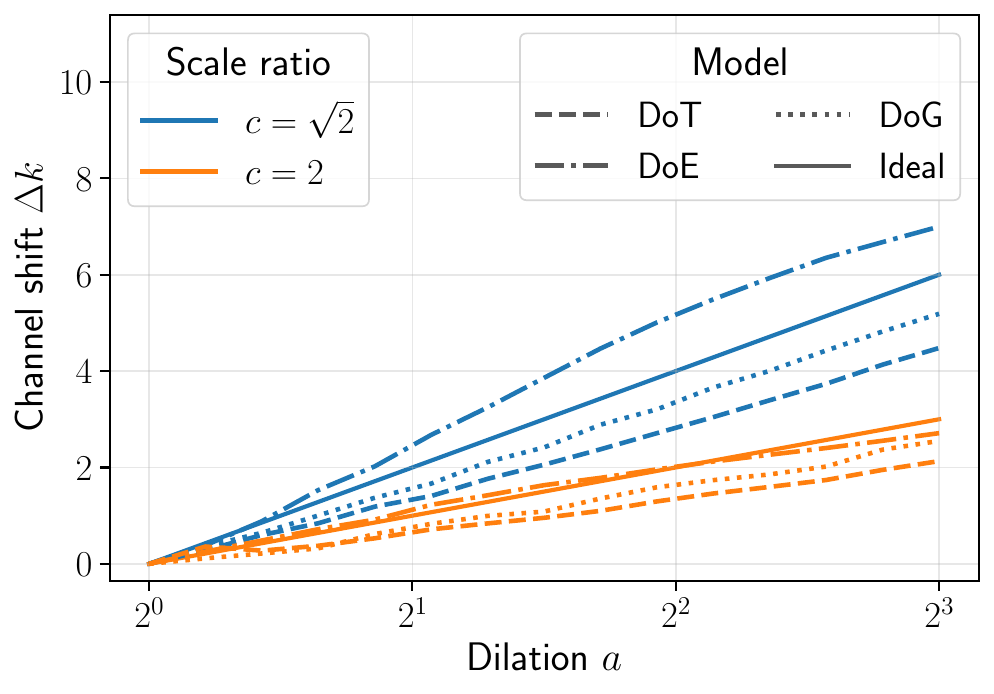}
    \caption{Scale covariance under temporal rescaling.
        By dilating the envelope \eqref{eq:gabor_probe} by a factor $a$ and encoded with the spiking wavelets, the spike-activity centroid migrates by $\Delta k$, as predicted by the covariance law $\log_c a$ (solid).
    }
    \label{fig:scale_covariance}
\end{figure}

\vspace{-0.4em}
\subsection{Scale-covariant encoding} \label{sec:scale_covariance_exp}
A scale-covariant encoder should shift the spike distribution across channels predictably when the input is dilated in time, where a non-covariant encoder would not.
To demonstrate this, we encode a sine wave with a Gaussian centered at $t_0$,
\begin{equation} \label{eq:gabor_probe}
    g(t; f_0) = \frac{1}{\|g\|_2} e^{\frac{-(t - t_0)^2}{2\DEV_g^2}} \sin(2\pi f_0 t), \quad \DEV_g = n_{\rm cyc}/f_0,
\end{equation}
dilate it by factors $a \in [1, 8]$, and measure the displacement of the per-channel spike-activity centroid.
Here $\DEV_g$ is the standard deviation of the Gaussian envelope and $n_{\rm cyc}$ the number of oscillations per $\DEV_g$, so every frequency carries the same number of oscillations and stretching the packet in time returns the same shape at $f_0/a$.
The packet is normalized to unit energy so that spike counts stay comparable across $a$; the centroid is the spike-count-weighted average across channels.
See \cref{sec:experimental_details} for $n_{\rm cyc}$ and the frequencies $f_0$.

\cref{fig:scale_covariance} shows that the measured shift follows the covariance law $\Delta k = \log_c a$: the two scale ratios separate closely along the expected slope, with finer spacing $c = \sqrt{2}$ shifting the activities by twice as many channels as $c=2$.
All three spiking wavelet families are covariant to within one channel at $c=2$ and $1.5$ at $c=\sqrt{2}$, and the residual grows at the largest $a$ as the activity centroid approaches the coarsest channel and can migrate no further.

\section{Discussion} \label{sec:discussion}
We contributed two time-causal and scale-covariant families of spiking wavelets as sparse continuous-time encoders.
We derived closed-form reconstruction error bounds for the spiking DoE and DoT wavelets and demonstrated matching or exceeding performance compared to non-causal Morlet wavelet at $c=2$ and the causal Szu wavelet, while remaining causal and sparse per channel.

A strength of this construction is its generality: the frame bounds and reconstruction guarantees hold for any finite-energy input, with no assumptions on signal statistics or stationarity.
The spiking reconstruction bound \eqref{eq:total_error_bound} additionally requires each channel to have an effective bandwidth, which the DoT cascade guarantees through its $O(|\omega|^{-N})$ decay.
The slower $O(|\omega|^{-1})$ decay of the DoE requires further assumptions on the signal spectrum.
The guarantees presented here are deterministic and worst-case, complementing an information-theoretic, average-case account of spike coding \cite{warland1996spikes, schneidman2006weak}.

Continuous-time spiking dynamics are simulated on a digital substrate at fixed time intervals.
The method's natural home is event-driven neuromorphic hardware, where continuous dynamics are native and the discretization artifacts disappear.
Our study establishes the feasibility of spiking wavelets for signal encoding and reconstruction, with a least-squares decoder that sets the upper bound any causal decoder must approach.
The baseline comparisons to $\Sigma\Delta$ and MC-IFTEM rely on the same decoder for fairness, but no deployable spike-only decoder is evaluated.
Closing the gap to online, streaming decoders is now a well-posed follow-up, as is quantifying the energy, latency, and memory benefits on analog or mixed-signal platforms.

Because every operation in the spiking wavelets reduces to a leaky-integrator difference with spike thresholding, the encoders map directly onto primitives common to neuromorphic platforms, positioning spiking wavelets as a principled building block for neuromorphic analog-to-digital conversion.
The wavelets encode time-varying signals into spike trains that are sparse within each of the $2(K{+}1)$ polarity channels, with provable reconstruction guarantees.
A single threshold $\theta_\mathrm{thr}$ sets the event rate, and the reconstruction error grows linearly with it \eqref{eq:limit-error-bound}.
Based on this theory and experiments, we propose the presented spiking DoE and DoT wavelets as fundamental spiking representations for neuromorphic processing.

% \section*{Acknowledgment}
% Support from the Novo Nordisk Foundation (NNF24OC0089302) and the Swedish Research Council (2022-02969) is gratefully acknowledged.

\bibliography{RefSNN}%,references}
\bibliographystyle{IEEEtran}

%%%%%%%%%%%%%%%%%%%%%%%%%%%%%%%%%%%%%%%%%%%%%%%%%%%%%%%%%%%%%%%%%%%%%%%%%%%%%%%
%%%%%%%%%%%%%%%%%%%%%%%%%%%%%%%%%%%%%%%%%%%%%%%%%%%%%%%%%%%%%%%%%%%%%%%%%%%%%%%
% APPENDIX
%%%%%%%%%%%%%%%%%%%%%%%%%%%%%%%%%%%%%%%%%%%%%%%%%%%%%%%%%%%%%%%%%%%%%%%%%%%%%%%
%%%%%%%%%%%%%%%%%%%%%%%%%%%%%%%%%%%%%%%%%%%%%%%%%%%%%%%%%%%%%%%%%%%%%%%%%%%%%%%

%\pagebreak

\appendices
% Appendix sections are still \section, so tell cleveref to name them "Appendix"
\crefalias{section}{appendix}
\crefalias{subsection}{appendix}
\crefalias{subsubsection}{appendix}

%%%%%%%%%%%%%%%%%%%%%%%%%%%%%%%%%%%%%%%%%%%%%%%%%%
\section{Neuronal implementations of scale-space wavelets} \label{app:sec:neuronal_implementation}

\subsection{Reconstruction constants}
To properly reconstruct the original signal, the contribution of each spiking channel needs to compensate for the scaling terms in both the difference kernels $\psi_{\rm DoE}$ and $\psi_{\rm DoT}$ and the LIF model.
Below, we first reverse the scaling from the coupled integration in the wavelet and LIF model.
We then derive the initialization scheme for the channel-specific weights based on naive assumptions on the input.

\subsubsection{Convolution of two truncated exponentials}
The spikes generated by the LIF are derived from the combination of two filters with two different temporal scale parameters: $\TIMECONSTANT_k$ for the $\psi_{\rm DoE}$ with $N=1$ \eqref{eq:doe_wavelet} or $\DEV_m$ (to first approximation) for the $\psi_{\rm DoT}$ \eqref{eq:time-causal_limit_kernel_convolution} (with a scale factor of exactly $\prod^\infty_{n=k} 1/\mu_n$ for the $k$th time-causal limit kernel) on the one hand, and $\mu_m$ for the LIF model on the other.
The composition of two truncated exponentials $h_1(t;\; \TIMECONSTANT_1) * h_2(t;\; \TIMECONSTANT_2)$ does not reduce to a single truncated exponential, but takes the form
\begin{align} \label{eq:composed_exponentials}
     & \int_{-\infty}^\infty h_1(t-\xi;\; \mu_1) \, h_2(\xi;\; \mu_2) \, {\rm d }\xi
    =\frac{1}{\mu_1 \, \mu_2}\int_0^t e^{-\frac{t-\xi}{\mu_1}}e^{-\frac{\xi}{\mu_2}} \, {\rm d }\xi
    \nonumber                                                                                                                                                 \\
     & =\frac{1}{\mu_1 \, \mu_2}e^{-\frac{t}{\mu_1}}\int_0^t e^{\xi(\frac{1}{\mu_2}-\frac{1}{\mu_1})} \, {\rm d }\xi \nonumber                                \\
     & =\frac{-\mu_1 \, \mu_2}{\mu_1 -\mu_2}\frac{1}{\mu_1 \, \mu_2}e^{-\frac{t}{\mu_1}} \left[e^{-\xi\frac{\mu_1 -\mu_2}{\mu_1 \mu_2}} \right]_0^t \nonumber \\
     & =\frac{1}{\mu_1 -\mu_2}\left[e^{-\frac{t}{\mu_1}}-e^{-\frac{t}{\mu_2}} \right].
\end{align}
The factor $1/(\mu_1 - \mu_2)$ is used to recover the scaling that occurs by the composition of the wavelet kernel and the leaky integrator term of the LIF neuron that produces the spikes.

\subsubsection{Weight initializations} \label{app:sec:weight_initialization}
To recover the original signal amplitude, we must recover the original integration value that led to a spike.
Recall that the bandpass formulation is based on the difference of exponentials as the generating function. We use $C_\kappa$ as a normalizing factor
\begin{equation}
    \kappa(t;\ \mu_1, \mu_2) = (\mu_1^{-1} e^{-t / \mu_1} - \mu_2^{-1} e^{-t / \mu_2}) / C_\kappa.
\end{equation}
Since $\int_0^\infty 1 / \mu \, e^{-t / \mu} = 1$, we have that
\begin{equation}
    C_\kappa = 1 / \mu_1 - 1 / \mu_2 = \frac{\mu_2 - \mu_1}{\mu_1 \, \mu_2}.
\end{equation}
We cannot infer the shape of the input signal that caused the spike, since there are infinitely many combinations of the leaky integrator and the input that integrates to $\theta_{\rm thr}$.
We can only know that the integration of the difference of the truncated exponentials, beginning at time $t_0$, reaches $\theta_{\rm thr}$ at time $t_1$,
\begin{align}
     & \theta_{\rm thr} = \int_{t_0}^{t_1} \kappa(t_1 - t;\; \mu_1, \mu_2) f(t)\ {\rm d}t \\
     & = \frac{\mu_2 - \mu_1}{\mu_1 \, \mu_2}\! \int_{t_0}^{t_1}\!\!
    \big( \mu_1^{-1} e^{-(t_1 - t) / \mu_1} -\mu_2^{-1} e^{-(t_1 - t) / \mu_2} \big) \ f(t)\ {\rm d}t. \nonumber
\end{align}
Assuming that $f(t)=f_1$ is constant over the interval and setting $w = (t_1 - t)$, which is 0 when $t = t_1$ so that ${\rm d}t = -{\rm d}w$, we can deduce $f_1$ from the interval $t_1 - t_0$, $\mu_1$, and $\mu_2$
\begin{align}
    \theta_{\rm thr} & = f_1 C_\kappa^{-1} \int_{t_0}^{t_1} \mu_1^{-1}e^{-(t_1 - t)/\mu_1} - \mu_2^{-1}e^{-(t_1 - t) / \mu_2} \ {\rm d}t \nonumber \\
                     & = f_1 C_\kappa^{-1} \int_{t_1 - t_0}^0 \mu_1^{-1}e^{-w / \mu_1} - \mu_2^{-1} e^{-w / \mu_2}\ (- {\rm d}w) \nonumber         \\
                     & = f_1 C_\kappa^{-1} \left[ - e^{-w/ \mu_1} - (- e^{-w / \mu_2}) \right]_0^{t_1 - t_0} \nonumber                             \\
                     & = f_1 C_\kappa^{-1}\left(e^{-(t_1 - t_0) / \mu_2} - e^{-(t_1 - t_0) / \mu_1} \right) \iff \nonumber                         \\
    f_1              & = \theta_{\rm thr}\ \frac{\mu_2 - \mu_1}{\mu_1 \, \mu_2} \frac{1}{e^{-(t_1 - t_0) / \mu_2} - e^{-(t_1 - t_0) / \mu_1}}.
    \label{eq:reconstruction_weights_default}
\end{align}

%%%%%%%%%%%%%%%%%%%%%%%%%%%%%%%%%%%%%%%%%%%%%%%%%%%%%%%%%%%%%%%%%%%%%%%%%%%%%%%%%%%
\end{document}